\documentclass{article}

\PassOptionsToPackage{numbers, compress}{natbib}
\usepackage[preprint]{neurips_2026}

\usepackage[utf8]{inputenc} %
\usepackage[T1]{fontenc}
\usepackage[hidelinks]{hyperref}       %
\usepackage{url}            %
\usepackage{booktabs}       %
\usepackage{amsfonts}       %
\usepackage{nicefrac}       %
\usepackage{microtype}      %
\usepackage{xcolor}         %
\usepackage{amsmath}
\usepackage{graphicx} 
\usepackage{multirow}
\usepackage{comment}
\usepackage{wrapfig}
 \usepackage{float}
\usepackage{subcaption}

\definecolor{datasetgreen}{RGB}{27,94,32}

\title{Scaling Optimization-Oriented Hypernetworks for Implicit Neural Representations}

\author{%
{\textbf{Lulu Cai\textsuperscript{1} \quad Hanxiang Ren\textsuperscript{2,3} \quad Siyan Dong\textsuperscript{1} \quad Li Sun\textsuperscript{1}}}\\
{\textbf{Jiefeng Wu\textsuperscript{1} \quad Youyi Zheng\textsuperscript{3} \quad Yi Ma\textsuperscript{1} \quad Yanchao Yang\textsuperscript{1}}}\\
{\textsuperscript{1}The University of Hong Kong 
\qquad \textsuperscript{2}TranscEngram
\qquad \textsuperscript{3}Zhejiang University}\\
{\small\texttt{\{lulu01cai,sunlids,jiefengwu\}@connect.hku.hk}}\\
{\small\texttt{\{siyan3d,mayi,yanchaoy\}@hku.hk} \quad \texttt{\{hanxiang.ren,youyizheng\}@zju.edu.cn}}}

\begin{document}

\maketitle

\begin{abstract}
We present a hypernetwork that generates the parameters of an implicit neural representation (INR) for a given signal in a single forward pass, without test-time optimization. Existing hypernetworks for this problem either ignore the chain-rule structure of the test-time optimization that produces an INR or, when they preserve it, instantiate the structure with MLPs whose parameter count grows quartically in target-network width. Our central contribution is to show that this parameter-count cost is structural to the MLP instantiation, not to the optimization-oriented principle of preserving chain-rule structure, and to resolve it by replacing the MLP body with attention. The resulting hypernetwork combines three design choices: an output-neuron-centric tokenization that aligns the token axis with each layer's output-neuron axis; an information-flow consistency rule that fixes the role of every cross-attention module from the representation its output should occupy; and a token-space iterative rollout that realizes the five quantities one optimization step on the target INR computes as five cross-attention modules with parameter count independent of target-network width. Across five 2D and 3D INR generation benchmarks, our method outperforms the state-of-the-art single-pass and optimization-oriented baselines; scales to target-network widths at which the MLP-based optimization-oriented paradigm exhausts GPU memory; and produces the highest result on NeRF generation without ground-truth INR weights supervision.
\end{abstract}

\vspace{-3pt}
\section{Introduction}
\label{sec:intro}
\vspace{-2pt}

Implicit neural representations (INRs) 
parameterize a signal 
as a neural network 
mapping coordinates to signal values, 
offering a continuous, 
resolution-independent, 
modality-agnostic functional view of data~\citep{sitzmann2019siren,tancik2020fourfeat,mildenhall2020nerf}.
The price of this functional view is computational: 
producing the INR for a single new signal requires a per-instance optimization
that takes thousands of gradient steps, 
shares no computation across signals, 
and is ill-suited to deployment scenarios 
in which
signals arrive at inference time. 
Eliminating this per-instance
cost -- 
generating an INR for a given signal in a single forward pass, without test-time optimization -- 
falls in the regime of hypernetworks.

However, 
current hypernetworks  
on this problem 
splits along an unresolved tradeoff. 
\emph{Single-pass} hypernetworks regress the
INR parameters directly~\citep{Chen2022TransformersAM,peis2025hypertransforming,koyuncu2023variational,Erkoc_2023_ICCV};
they scale to large target networks 
and large training sets but
treat the per-layer weights as a flat output, 
discarding the chain-rule structure that ties those weights together in the optimization that produced them. 
\emph{Optimization-oriented}
hypernetworks instead unroll an optimization procedure into the architecture, 
with HyPoGen~\citep{ren2025hypogen} predicting
parameter updates layer by layer in the same forward-then-backward order that gradient descent on the target INR would follow,
preserving the inter-layer dependencies of $\partial \mathcal{L} /
\partial W_i$ 
that single-pass methods discard. 
Yet, HyPoGen's MLP-based instantiation 
incurs a parameter cost quartic
in target-network width 
per refinement step, 
leaving the larger INRs 
at which amortization is most useful out of reach.

Our position is that 
this tradeoff is structural 
to the MLP instantiation, 
not to the optimization-oriented principle. 
The factor that makes the MLP cost quartic 
is the joint dependence on
both the row count and the row dimension of each weight matrix;
attention removes the row-count dependence by sharing parameters across the token axis, 
and the row-dimension dependence is absorbed into a fixed embedding dimension by tokenization. 
The remainder of the paper 
develops the resulting hypernetwork: 
each weight matrix is tokenized output-neuron 
by output-neuron, 
the five quantities a single optimization step 
on the target INR computes -- one activation, two local Jacobians, the propagated loss-signal, and the parameter gradient -- 
are realized one-for-one as five cross-attention modules in a token-space iterative module,
and the QKV assignment of every module is fixed by a single design rule we call \emph{information-flow consistency}. 
The five modules together rollout for $T$ refinement steps, 
mirroring $T$ optimization steps on the target INR (Sec.~\ref{sec:method}).

In summary, 
we make three contributions: 
(i) we identify the scalability barrier of optimization-oriented hypernetworks 
as a structural property of the MLP instantiation rather than of the principle,
and remove it by replacing MLP sub-modules with attention; 
(ii) we derive a token-space realization of one optimization step in which five cross-attention modules -- tokenization, role assignments, and
all -- follow from a single design rule; 
(iii) on five INR generation benchmarks, 
our method outperforms the strongest representatives of both prior paradigms, scales to target-network widths at which the MLP-based optimization-oriented paradigm exhausts GPU memory, 
and produces the highest result 
on NeRF generation 
without resorting to weight supervision 
-- 
the empirical signature of the chain-rule structure that single-pass methods lack the architectural mechanism to enforce.

\vspace{-3pt}
\section{Related Works}
\label{sec:related}
\vspace{-2pt}

\vspace{-1pt}
\paragraph{Implicit neural representations (INRs).}
INRs encode signals as 
continuous functions that map 
coordinates to signal values, 
parameterized by neural networks~\citep{sitzmann2019siren,tancik2020fourfeat,mildenhall2020nerf}.
They underpin a range of applications including 
images and signed-distance fields~\citep{Park_2019_CVPR,Atzmon_2020_CVPR,chen2018implicit_decoder},
occupancy-based 3D shape reconstruction~\citep{Mescheder2019OccupancyNetworks,Genova2020LocalDeepImplicit,chibane20ifnet},
and radiance fields~\citep{mildenhall2020nerf,yu2021pixelnerf,Wang2021IBRNet,chen2021mvsnerf}.
Subsequent work has improved their expressivity through periodic, wavelet, and learned activations~\citep{sitzmann2019siren,saragadam2023wire,Liu2024FINER,kazerouni2024incode,fathony2021multiplicative,shi2024improved,Gao2024HSIRENII,Zhao2024QuantumIN},
multi-scale and band-limited
encodings~\citep{tancik2020fourfeat,lindell2021bacon,Mehta2021ModulatedINR},
hash- and grid-based encodings~\citep{mueller2022instant,Chen2022ECCV,takikawa2021nglod},
and transferable initializations~\citep{Tancik2021LearnedInit,vyas2024learning,xie2023diner}.
These advances enrich what an INR can represent but leave the per-instance test-time optimization untouched; amortizing it across a training distribution is the focus of hypernetworks.

\vspace{-1pt}
\paragraph{Hypernetworks for INR generation.}
A hypernetwork~\citep{ha2017hypernetworks} produces target-network parameters from a conditioning input, 
with utility demonstrated across audio~\citep{szatkowski2022hypersound},
parameter-efficient transfer~\citep{karimi2021parameterefficient,ortizbarajas2024hyperloaderintegratinghypernetworkbasedlora,ivison-peters-2022-hyperdecoders},
continual~\citep{ohs2019hypercl} and few-shot learning~\citep{sendera2023hypershot,Przewiezlikowski2022HyperMAMLFA},
tabular foundation models~\citep{mueller2025mothernet}, 
and weight prediction for unseen architectures~\citep{knyazev2021parameter,knyazev2023canwescale,Zhmoginov2022HyperTransformer}.
Within INR generation, 
the single-pass paradigm spans Transformer-based set-to-set or latent-to-weight mappings~\citep{Chen2022TransformersAM,peis2025hypertransforming},
low-dimensional modulation~\citep{Kim2022GeneralizableIN}, 
INR-VAE mixtures~\citep{koyuncu2023variational}, 
architectural extensions~\citep{babuhyperfields,hedlin2025hypernet,yang2023neumeta,wu2023hyperinr},
and a parallel line of weight-space generative models~\citep{Erkoc_2023_ICCV,wang2024neural,soro2025diffusionbased,schurholt2022hyperrepresentations,schuerholt2024sane,wang2026scaling,bedionita2025instructionguidedautoregressiveneuralnetwork}.
NeRF-specific hypernetworks~\citep{gu2023generalizable,lee2023localityaware,gu2025foundation} adapt the single-pass paradigm to radiance-field generation through domain-specific architectural choices such as locality-aware modulation and foundation-model conditioning.
The optimization-oriented paradigm is represented by HyPoGen~\citep{ren2025hypogen}, which unrolls a finite-step optimization procedure into the architecture and 
predicts per-step parameter updates whose layer-wise structure mirrors the chain-rule computation that gradient descent on the target INR would perform. 
We adopt this principle but observe that its MLP-based instantiation incurs a parameter cost quartic in target-network width per refinement step, 
making it prohibitive at the INR scales. 
Our method, 
to our knowledge, 
is the first to preserve the optimization-oriented principle while realizing it with attention, retaining the chain-rule structure that single-pass methods discard while overcoming the scalability barrier that confined this paradigm to target networks.

\vspace{-3pt}
\section{Method}
\label{sec:method}
\vspace{-2pt}

\vspace{-2pt}
\subsection{From Test-Time Optimization to Feed-Forward Generation}
\label{sec:method:goal}
\vspace{-1pt}

\vspace{-1pt}
\paragraph{Test-time optimization.}
An implicit neural representation (INR) encodes a signal $\zeta$ -- such as an image, a 3D shape, or a radiance field -- as a neural network $f_\theta: 
\mathbb{R}^{d_{\mathrm{in}}} \to \mathbb{R}^{d_{\mathrm{out}}}$ that maps a coordinate $\mathbf{x}$ to the signal value at that coordinate. 
The conventional way to obtain the parameters $\theta_\zeta$ of an INR for a given signal $\zeta$ is to solve a per-instance optimization problem,
\begin{equation}
\label{eq:method-per-instance}
\theta_\zeta \;=\; \arg\min_\theta \mathcal{L}(f_\theta, \zeta),
\end{equation}
where $\mathcal{L}$ is a reconstruction loss between $f_\theta$ and $\zeta$. 
We refer to this procedure as \emph{test-time optimization}, 
since the optimization is performed at the time the representation is needed, on the specific signal at hand, which 
is expensive on three counts: 
each signal incurs its own optimization run that takes minutes to hours, 
no computation is shared across signals,
and the procedure is unsuited to deployment scenarios in which signals arrive at inference time.

\vspace{-1pt}
\paragraph{Feed-forward generation and HyPoGen.}
A feed-forward alternative removes test-time optimization entirely: 
a hypernetwork $\mathcal{H}_\Omega$, trained once, produces the INR parameters in a single forward pass,
$\theta_\zeta = \mathcal{H}_\Omega(\mathbf{z}_\zeta)$, 
given a per-signal conditioning input $\mathbf{z}_\zeta \in \mathbb{R}^{d_z}$. 
The obstacle is that the per-layer weights
$W_1, \dots, W_L$ of $f_\theta$ are not independent: 
they are coupled through the chain-rule computation that test-time optimization would normally exhibit,
and a hypernetwork that ignores this coupling and regresses $\theta_\zeta$ as a flat output vector forfeits the structure
that makes the parameters consistent with one another.
\citep{ren2025hypogen} (HyPoGen) addressed this obstacle by \emph{unrolling} the test-time optimization procedure into the hypernetwork's architecture: 
rather than regressing $\theta_\zeta$ in one shot, 
HyPoGen produces it through $T$ feed-forward steps,
\begin{equation}
\label{eq:method-hypogen-iter}
\theta^{t+1} \;=\; \theta^{t} + \lambda^{t} \, \Delta\theta^{t},
\qquad t = 0, 1, \dots, T-1,
\end{equation}
where the step size $\lambda^{t} \in \mathbb{R}$ and 
the update direction $\Delta\theta^{t}$ are predicted by sub-modules of the hypernetwork, 
with the per-layer increment at layer $i$ depending on signals from all subsequent layers $j > i$ to mirror the chain-rule computation of $\partial \mathcal{L} / \partial W_i$.
We refer to this design 
as the \emph{optimization-oriented} principle and adopt it as our starting point.

\begin{figure}[t]
\centering
\includegraphics[width=0.95\linewidth]{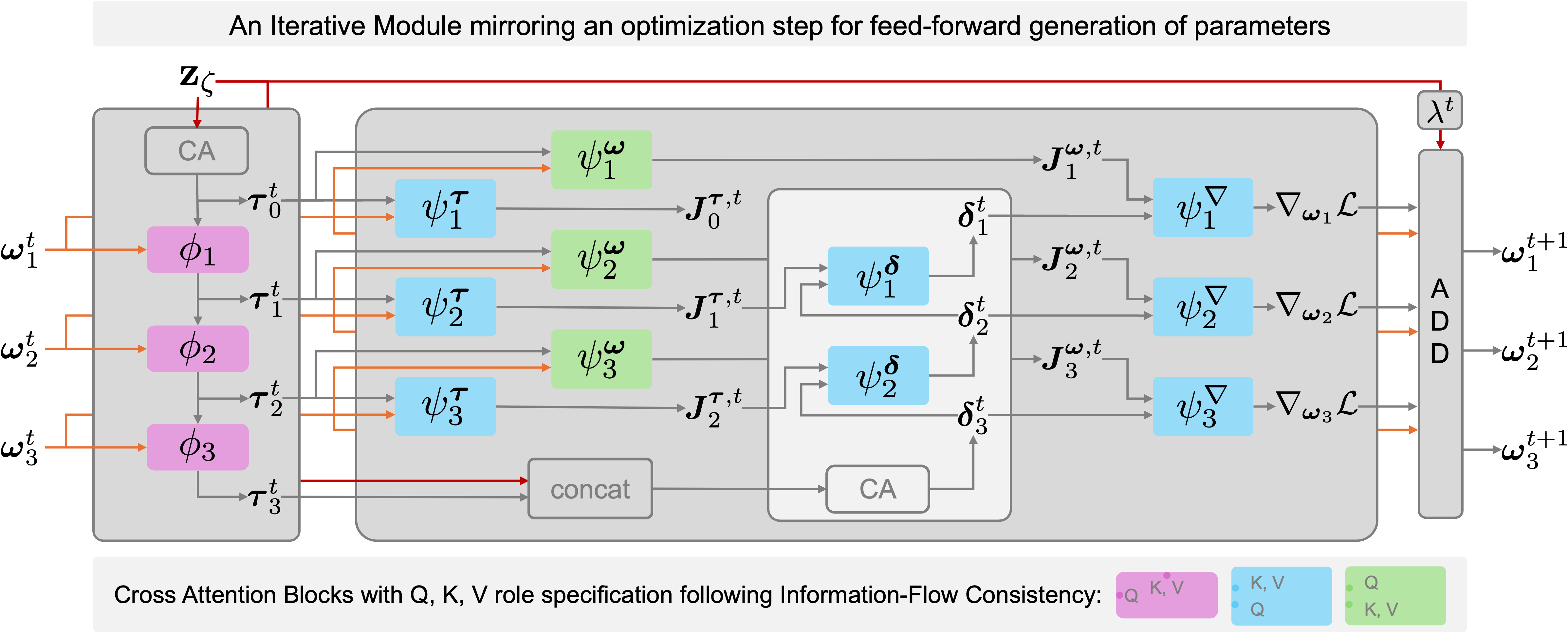}
\caption{
One iterative module 
of the proposed hypernetwork
$\mathcal{H}_\Omega$, 
illustrated for a 
three-layer MLP target
INR ($L=3$); 
the construction extends to 
arbitrary $L$. 
At iteration $t$, 
the forward chain (left, modules $\phi_i$)
produces the pseudo-activation tokens $\boldsymbol{\tau}_i^t$
from the parameter tokens $\boldsymbol{\omega}_i^{(t)}$, 
seeded by $\boldsymbol{\tau}_0^t$ 
from $\mathbf{z}_\zeta$ through the
Cross-Attention (CA) block. 
The backward block (right) 
realizes one chain-rule step 
at each layer: 
$\psi^{\boldsymbol{\tau}}_i$
and $\psi^{\boldsymbol{\omega}}_i$ 
produce the local Jacobian tokens $\boldsymbol{J}^{\boldsymbol{\tau},t}_{i-1}$ 
and
$\boldsymbol{J}^{\boldsymbol{\omega},t}_i$, which
$\psi^{\boldsymbol{\delta}}_{i-1}$ and $\psi^{\nabla}_i$
contract with $\boldsymbol{\delta}_i^t$ 
to produce the propagated loss-signal $\boldsymbol{\delta}_{i-1}^t$ and the parameter-gradient tokens
$\nabla_{\boldsymbol{\omega}_i}\mathcal{L}$;
the backward recursion is seeded by $\boldsymbol{\delta}_L^t$, 
formed from $\boldsymbol{\tau}_L^t$ 
and $\mathbf{z}_\zeta$ via the
concat-plus-CA module. 
The ADD block then applies the refinement
with predicted step size 
and forwards to the next module.
}
\vspace{-3pt}
\label{fig:iterative_module}
\end{figure}

\vspace{-1pt}
\paragraph{Scalability barrier of HyPoGen.}
The optimization-oriented principle is sound, but its instantiation in HyPoGen does not scale to the target networks of interest in INR generation. 
INRs are most commonly realized as MLPs with weight matrices $W_i \in \mathbb{R}^{n_i \times n_{i-1}}$, and HyPoGen's hypernetwork is itself MLP-based: 
each refinement step in Eq.~\ref{eq:method-hypogen-iter} produces $\Delta\theta^{(t)}$ through MLP sub-modules whose parameter count per layer per refinement step is $\mathcal{O}(n_i^2 \cdot n_{i-1}^2)$, quartic in target-network width. 
Compressing each row of $W_i$ to a $d$-dimensional latent before the MLP body reduces the per-row factor from $n_{i-1}^2$ to $d^2$ but leaves the $n_i^2$ factor intact; 
the full comparison is in
Appendix~\ref{app:param-count}. 
For large INRs and datasets 
that make hypernetwork-based amortization worthwhile, 
this cost is prohibitive.
Preserving the principle while overcoming the scalability barrier requires an architectural ingredient whose parameter count is decoupled from the dimensions of the target-network layer, and attention is such an ingredient: 
a cross-attention module operates on a token sequence through projections whose size depends only on the per-token dimension.
The remainder of the section develops a hypernetwork that retains the optimization-oriented principle and replaces the MLP body with attention, with the overall structure of one refinement step illustrated in Fig.~\ref{fig:iterative_module}.

\vspace{-2pt}
\subsection{What One Optimization Step Must Compute}
\label{sec:method:derivation}
\vspace{-1pt}

By the optimization-oriented principle, 
the proposed hypernetwork is structured as a rollout of an iterative parameter-update procedure, 
in which each iterative module mirrors the operations of a single optimization step on the target INR. 
Before specifying the neural modules,
we identify the quantities a single optimization step must compute in the target-network's own parameter and activation space.

\vspace{-1pt}
\paragraph{Per-instance loss and forward pass.}
Test-time optimization on a signal $\zeta$ minimizes the reconstruction loss $\mathcal{L}(f_\theta, \zeta)$ over the INR parameters $\theta = (W_1, \dots, W_L)$ by gradient-based updates. 
The forward pass of $f_\theta$ proceeds layer by layer,
\begin{equation}
\label{eq:method-fwd}
\mathbf{h}_i \;=\; \sigma\big(W_i \, \mathbf{h}_{i-1}\big),
\qquad i = 1, \dots, L,
\end{equation}
with $\mathbf{h}_0 = \mathbf{x}$ the input coordinate and $\mathbf{h}_L = f_\theta(\mathbf{x})$ the predicted signal value. 
Bias terms are omitted for clarity; their inclusion does not affect the structure of the derivation.

\vspace{-1pt}
\paragraph{Loss-signal and chain rule.}
Define the \emph{loss-signal at layer $i$} as $\delta_i := \partial \mathcal{L} / \partial \mathbf{h}_i$, 
with base case $\delta_L = \partial \mathcal{L} / \partial \mathbf{h}_L$ at the output layer.
Applying the chain rule to Eq.~\ref{eq:method-fwd} 
yields the two combinations a single optimization step requires at layer
$i$,
\begin{equation}
\label{eq:method-chain}
\delta_{i-1} \;=\; \delta_i \cdot
\frac{\partial \mathbf{h}_i}{\partial \mathbf{h}_{i-1}},
\qquad\qquad
\nabla_{W_i} \mathcal{L} \;=\; \delta_i \cdot
\frac{\partial \mathbf{h}_i}{\partial W_i},
\end{equation}
the first propagating the loss-signal one layer backward and
the second producing the parameter gradient that a gradient-based optimizer would use to update $W_i$.

\vspace{-1pt}
\paragraph{Five per-layer quantities.}
Reading off Eqs.~\ref{eq:method-fwd}--\ref{eq:method-chain}, 
one optimization step on the target INR computes, at each layer $i$, 
five quantities: 
(i) the activation $\mathbf{h}_i$; 
(ii) the activation Jacobian $\partial \mathbf{h}_i / \partial \mathbf{h}_{i-1}$; 
(iii) the parameter Jacobian $\partial \mathbf{h}_i / \partial W_i$; 
(iv) the propagated loss-signal $\delta_{i-1}$; 
and (v) the parameter gradient $\nabla_{W_i}\mathcal{L}$. 
Quantities (i)--(iii) are local,
each depending only on the current layer's parameters and the previous layer's activation; 
quantities (iv) and (v) are non-local, since they consume $\delta_i$, which depends on every layer downstream of $i$ through Eq.~\ref{eq:method-chain}.
The five cross-attention modules of the iterative module
(Fig.~\ref{fig:iterative_module}) realize these five quantities
in the token space. 
Next, 
we introduce the parameter tokens $\boldsymbol{\omega}_i$ that play the role
of $W_i$ in this token-space machinery.

\vspace{-2pt}
\subsection{Output-Neuron-Centric Tokenization}
\label{sec:method:tokenization}
\vspace{-1pt}

\begin{figure}[!t]
\centering
\includegraphics[width=0.95\linewidth]{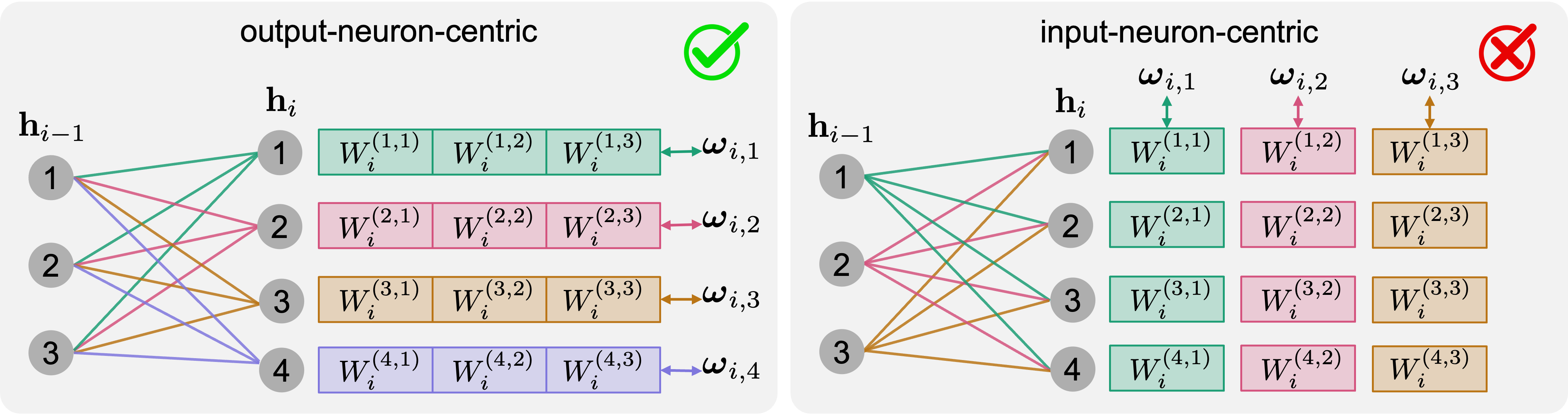}
\caption{Two tokenization schemes 
for the layer-$i$ weight matrix $W_i$.
\textbf{Left (ours):} 
the \emph{output-neuron-centric} scheme
groups each row of $W_i$ into one token $\boldsymbol{\omega}_{i,j}$, 
placing each output neuron's parameters into a single, co-adapted unit. 
\textbf{Right:} the \emph{input-neuron-centric} scheme groups each column of $W_i$ into one token, 
collecting parameters that originate from one input neuron to multiple output neurons that may encode unrelated features and are therefore not co-adapted. 
}
\vspace{-3pt}
\label{fig:tokenization}
\end{figure}

To replace HyPoGen's MLP body with attention, 
we must choose how to tokenize the weight matrix $W_i \in \mathbb{R}^{n_i \times n_{i-1}}$ of each target-network layer. 
Two natural schemes are illustrated in Fig.~\ref{fig:tokenization}: 
the \emph{output-neuron-centric} scheme that slices $W_i$ row by row, 
producing a token sequence of length $n_i$, 
and the \emph{input-neuron-centric} scheme that slices $W_i$ column by column, 
producing a token sequence of length $n_{i-1}$. 
Both fit the input format that attention requires, but the two schemes group different weight entries into one token, and the choice of which entries to group is what makes one scheme suit the hypernetwork's task while the other does not.

\vspace{-1pt}
\paragraph{Tokenization.}
We adopt the output-neuron-centric scheme and represent $W_i$ as
\begin{equation}
\label{eq:method-omega-tokens}
\boldsymbol{\omega}_i \;=\; \big(\boldsymbol{\omega}_{i,1};
\dots; \boldsymbol{\omega}_{i,n_i}\big) \;\in\; \mathbb{R}^{n_i
\times d},
\end{equation}
where the $j$-th token $\boldsymbol{\omega}_{i,j} \in \mathbb{R}^{1 \times d}$ is the encoding of the $j$-th row $W_i^{(j,:)} \in \mathbb{R}^{n_{i-1}}$ through a learned per-layer encoder, and a per-layer decoder $\mathrm{Decode}_i:
\mathbb{R}^d \to \mathbb{R}^{n_{i-1}}$ recovers the row when the actual weight is needed.
The token sequence $\boldsymbol{\omega}_i$ is the hypernetwork-side analogue of
$W_i$ named above.

\vspace{-1pt}
\paragraph{Why row-wise rather than column-wise.}
The forward pass of the target INR (Eq.~\ref{eq:method-fwd}) is row-wise: 
the $j$-th component of $\mathbf{h}_i$ is determined entirely by the $j$-th row of $W_i$ acting on the entire previous activation $\mathbf{h}_{i-1}$. 
The $n_{i-1}$ entries of one row therefore form a co-adapted group, 
in the sense that they jointly determine one output neuron and their values must be consistent with one another for that neuron to compute the function the network needs. 
A column of $W_i$, by contrast, collects parameters that fan out from one input neuron to $n_i$ different output neurons that may encode unrelated features; 
the entries of a column are not co-adapted, and compressing them into a single token offers no functional unit for the attention mechanism to attend and the decoder to recover.

\vspace{-1pt}
\paragraph{Token-axis as output-neuron-axis.}
Since the $j$-th row of $W_i$ is exactly the parameters that produce the $j$-th output neuron, each token $\boldsymbol{\omega}_{i,j}$ corresponds to one output neuron of layer $i$. 
All five per-layer quantities of Sec.~\ref{sec:method:derivation} live on this axis: 
$\mathbf{h}_i$ has one entry per output neuron, 
the activation Jacobian
$\partial \mathbf{h}_i / \partial \mathbf{h}_{i-1}$ 
has one row per output neuron, 
the parameter Jacobian $\partial \mathbf{h}_i / \partial W_i$ 
is block-diagonal across output neurons (since $h_i^{(j)}$ depends only on the $j$-th row of $W_i$), 
and both $\delta_i$ and $\nabla_{W_i}\mathcal{L}$ have one entry or row per output neuron. 
Under output-neuron-centric tokenization, 
the token-space analogues of these quantities share the token axis with $\boldsymbol{\omega}_i$, so that attention's native operation of mixing across tokens corresponds to mixing across output neurons of the target layer. 
The empirical comparison against the
input-neuron-centric scheme is reported in Sec.~\ref{sec:analysis}.

\vspace{-2pt}
\subsection{The Iterative Module}
\label{sec:method:iterative-module}
\vspace{-1pt}

The iterative module (Fig.~\ref{fig:iterative_module}) produces the remaining five token-space quantities through five cross-attention modules: 
a forward chain of modules $\phi_i$ produces the pseudo-activation tokens $\boldsymbol{\tau}_i$, 
and a backward block of four modules $\psi^{\boldsymbol{\tau}}_i,
\psi^{\boldsymbol{\omega}}_i, \psi^{\boldsymbol{\delta}}_{i-1},
\psi^{\nabla}_i$ produces the activation-Jacobian tokens
$\boldsymbol{J}^{\boldsymbol{\tau}}_{i-1}$, 
the parameter-Jacobian tokens $\boldsymbol{J}^{\boldsymbol{\omega}}_i$, 
the loss-signal tokens $\boldsymbol{\delta}_{i-1}$, 
and the parameter-gradient tokens $\nabla_{\boldsymbol{\omega}_i}\mathcal{L}$.
The trainable parameters of the iterative module are the FFNs inside each cross-attention module; the QKV projections operate on tokens of dimension $d$ and contribute parameters at scale $\mathcal{O}(d^2)$ independent of $n_i$ and $n_{i-1}$.

\vspace{-1pt}
\paragraph{Information-flow consistency.}
We design every cross-attention module in the iterative module by a single rule. 
Cross-attention's output is a row-wise convex combination of the rows of $V$ and therefore inherits the representation of $V$, 
while the row count of the output is inherited from $Q$. 
The role of each module's output is fixed in advance -- 
the forward block must produce pseudo-activation tokens, the activation-Jacobian module must produce activation-Jacobian tokens, and so on -- 
so the slot $V$, and consequently $K$, must be occupied by the input that already lives in the role the output should occupy, with $Q$ the remaining input. 
We call this rule \emph{information-flow
consistency} and apply it uniformly to derive the QKV assignment of every module below; 
the resulting assignments are also encoded by the color coding in Fig.~\ref{fig:iterative_module}.

\vspace{-1pt}
\paragraph{Forward chain.}
The forward block at layer $i$ takes $\boldsymbol{\omega}_i$ and $\boldsymbol{\tau}_{i-1}$ as inputs and produces an output in the role of pseudo-activation tokens for the current layer.
By information-flow consistency, $\boldsymbol{\tau}_{i-1}$ is the input that already lives in the pseudo-activation role, 
so $V = K = \boldsymbol{\tau}_{i-1}$ and $Q = \boldsymbol{\omega}_i$, 
giving
\begin{equation}
\label{eq:method-forward}
\boldsymbol{\tau}_i \;=\; \phi_i(\boldsymbol{\omega}_i,
\boldsymbol{\tau}_{i-1}) \;=\;
\mathrm{softmax}\!\left(\frac{\boldsymbol{\omega}_i \,
\boldsymbol{\tau}_{i-1}^\top}{\sqrt{d}}\right)
\boldsymbol{\tau}_{i-1}.
\end{equation}
This module reproduces the structural skeleton of a standard INR layer (Eq.~\ref{eq:method-fwd}) component by component:
the score matrix plays the role of the linear mixing $W_i \mathbf{h}_{i-1}$, 
the row-wise softmax plays the role of the non-linearity $\sigma$, and the row-wise tokenization of Sec.~\ref{sec:method:tokenization} ensures output-unit alignment between the $j$-th row of $W_i$ and the $j$-th output token. 
Recursive application of $\phi_i$ for $i = 1, \dots, L$ mirrors the forward pass of Eq.~\ref{eq:method-fwd} in token space; 
the recursion is seeded by $\boldsymbol{\tau}_0$,
produced from $\mathbf{z}_\zeta$ through the top-left cross-attention module in Fig.~\ref{fig:iterative_module}. 
The full component-by-component correspondence is in Appendix~\ref{app:forward-correspondence}.

\vspace{-1pt}
\paragraph{Backward block.}
The backward block at layer $i$ realizes operations (ii)--(v) of Sec.~\ref{sec:method:derivation} as four cross-attention modules, 
with each module's QKV assignment fixed by information-flow consistency once the role of its output is read off the specification:
\begin{align}
\label{eq:method-Jtau}
\boldsymbol J^{\boldsymbol{\tau}}_{i-1} &=
\psi^{\boldsymbol\tau}_i(\boldsymbol{\omega}_i,
\boldsymbol{\tau}_{i-1}), &
\boldsymbol J^{\boldsymbol{\omega}}_i &=
\psi^{\boldsymbol\omega}_i(\boldsymbol{\tau}_{i-1},
\boldsymbol{\omega}_i), \\
\label{eq:method-delta-grad}
\boldsymbol{\delta}_{i-1} &=
\psi^{\boldsymbol\delta}_{i-1}(\boldsymbol{\delta}_{i},
\boldsymbol J^{\boldsymbol{\tau}}_{i-1}), &
{\nabla}_{\boldsymbol{\omega}_i}\mathcal{L} &=
\psi^{\nabla}_i(\boldsymbol{\delta}_i, \boldsymbol
J^{\boldsymbol{\omega}}_i),
\end{align}
where each module argument list 
is written in the order $(Q,
\text{shared } K\!=\!V)$. 
The recursion runs with $i$ ranging from $L$ down to $1$, 
each step producing
$\boldsymbol{\delta}_{i-1}$ 
from $\boldsymbol{\delta}_i$ and the
layer-$i$ Jacobian token sequences. 
The recursion is seeded by
$\boldsymbol{\delta}_L$, 
produced from $\boldsymbol{\tau}_L$ and
$\mathbf{z}_\zeta$ through the concat-plus-cross-attention module on the right of Fig.~\ref{fig:iterative_module}; 
the two inputs carry the prediction-side and signal-side information that the target-network loss-signal 
$\delta_L = \partial \mathcal{L} / \partial \mathbf{h}_L$ depends on. 
Per-module derivations of the QKV assignments 
following the {\it information-flow consistency} principle
are in Appendix~\ref{app:qkv-derivations}.

\vspace{-2pt}
\subsection{Iterative Refinement and the Full Hypernetwork}
\label{sec:method:rollout}
\vspace{-1pt}

The forward chain 
and the four backward modules  
furnish the per-layer machinery 
for one refinement step on the target INR. 
Now we compose the per-layer machinery 
into one refinement step 
of the target network 
and stack refinement steps 
across $T$ iterations 
to realize the full hypernetwork.

\vspace{-1pt}
\paragraph{One refinement step.}
One refinement step 
proceeds in three stages 
as shown in Fig.~\ref{fig:iterative_module}. 
The forward chain of $\phi_1, \dots, \phi_L$ is run across all $L$ 
target-network layers, 
producing the pseudo-activation tokens $\boldsymbol{\tau}_0, \dots, \boldsymbol{\tau}_L$. 
The backward block is then run with $i$ ranging from $L$ down to $1$,
producing the loss-signal tokens $\boldsymbol{\delta}_{L-1}, \dots, \boldsymbol{\delta}_0$ 
together with the parameter-gradient tokens
$\nabla_{\boldsymbol{\omega}_i}\mathcal{L}$ at each layer as side outputs. 
Each layer's parameter tokens are then refined by
\begin{equation}
\label{eq:method-refinement-step}
\boldsymbol{\omega}_i^{t+1} \;=\;
\boldsymbol{\omega}_i^{t} \;-\; \lambda^{t} \,
\nabla_{\boldsymbol{\omega}_i}\mathcal{L}^{t},
\qquad i = 1, \dots, L,
\end{equation}
where $\lambda^{t} \in \mathbb{R}$ 
is a step size predicted by a sub-module of the hypernetwork from $\mathbf{z}_\zeta$.
Eq.~\ref{eq:method-refinement-step} is one iteration of the rollout in Eq.~\ref{eq:method-hypogen-iter}, 
applied to the parameter token sequences directly; 
the equivalence with the weight-space rollout is up to the per-layer decoder
$\mathrm{Decode}_i$ of Sec.~\ref{sec:method:tokenization}.

\vspace{-1pt}
\paragraph{Rollout and the full hypernetwork.}
Repeating the iterative module 
for $T$ iterations 
produces parameter token sequences $\boldsymbol{\omega}_i^{0}, \dots, \boldsymbol{\omega}_i^{T}$ at each layer $i$. 
The initial sequence $\boldsymbol{\omega}_i^{0}$ is generated from $\mathbf{z}_\zeta$ through an initialization sub-module, 
and the final sequence $\boldsymbol{\omega}_i^{T}$ 
is decoded to the target-network weight matrix $W_i$. 
The full hypernetwork is the composition 
of these $T$ modules,
\begin{equation}
\label{eq:method-hypernetwork}
\theta_\zeta \;=\; \mathcal{H}_\Omega(\mathbf{z}_\zeta) \;=\;
\big(W_1, \dots, W_L\big), \qquad W_i \;=\;
\mathrm{Decode}_i\!\big(\boldsymbol{\omega}_i^{T}\big),
\end{equation}
producing the target-INR parameters in a single forward evaluation of $\mathcal{H}_\Omega$ as Sec.~\ref{sec:method:goal} required. 
The trainable parameters of $\mathcal{H}_\Omega$
comprise the per-layer encoders and decoders (Appendix~\ref{app:method:encode-decode}), the FFNs inside
the five cross-attention modules of the iterative module, 
the step-size sub-module, 
and the initialization sub-modules for
$\boldsymbol{\tau}_0$, $\boldsymbol{\delta}_L$, and
$\boldsymbol{\omega}_i^{0}$ at each layer; 
details of the initialization 
sub-modules are in Appendix~\ref{app:init}. 
The parameter count of the core iterative body is independent of the target-network width at each layer, 
with the full comparison against HyPoGen given in Appendix~\ref{app:param-count}.

\vspace{-3pt}
\section{Experiments}
\label{sec:experiments}
\vspace{-2pt}

We evaluate our method 
on a diverse set of INR generation tasks
covering both 2D and 3D signals, 
across five datasets. 
For 2D signals, 
we adopt CelebA-HQ~\citep{liu2015deep,karras2018progressive} 
and ImageNet~\citep{russakovsky2015imagenet} 
for natural image tasks,
and ERA5~\citep{hersbach2019era5} for climate data. 
For 3D signals,
we use ShapeNet Chairs~\citep{Chang2015ShapeNetAI} for geometric reconstruction and 
Pre-trained NeRF~\citep{Ramirez2023DeepLO} for
novel view synthesis. 
Major baselines are
HyPoGen~\citep{ren2025hypogen}, 
an MLP-based optimization-oriented hypernetwork, 
and LDMI~\citep{peis2025hypertransforming}, 
an attention-based hypernetwork without optimization-oriented design;
together they represent the two existing paradigms that our method unifies. 
These baselines isolate the two axes of our
contribution: 
HyPoGen instantiates the optimization-oriented
principle without attention, 
LDMI uses attention without the optimization-oriented principle, 
and our method combines the two.
We additionally include
VaMoH~\citep{koyuncu2023variational} 
on the 2D datasets and ShapeNet Chairs, 
and HyperDiffusion~\citep{Erkoc_2023_ICCV} on Pre-trained NeRF; 
Trans-INR~\citep{Chen2022TransformersAM} and
GINR-IPC~\citep{Kim2022GeneralizableIN} are deferred to Appendix~\ref{app:additional-quantitative-results}. 
For each task, 
a shared encoder across all methods, 
maps the signal to $\mathbf{z}_\zeta$, 
from which the hypernetwork 
generates the target INR parameters $\theta_\zeta$ (Appendix~\ref{App:encoder architectures}).
Architectures, training configurations, loss functions, additional qualitative results, 
and the LPIPS metric on ImageNet are in Appendix~\ref{app:Implementation Details-new}.

\vspace{-2pt}
\subsection{2D INR Generation}
\label{sec:2d_eval}
\vspace{-1pt}

For 2D signals, 
we report reconstruction quality
with PSNR, SSIM, and LPIPS in Table~\ref{tab:hyperinr-2d}.

\vspace{-1pt}
\textbf{CelebA-HQ}~\citep{liu2015deep,karras2018progressive} 
contains 30,000 face images (28K train, 2K test). 
Following~\citep{Dupont2022FromDT}
and~\citep{peis2025hypertransforming}, 
we use a resolution of $64{\times}64$. 
Our method outperforms all competitors 
across the three metrics, 
with a PSNR gain of 2.9\,dB over LDMI, 
and produces reconstructions that closely match the ground truth. 
The gap against HyPoGen is noticeable: 
the MLP body's quartic parameter growth (Sec.~\ref{sec:method:goal}) {\it does not} endow it the capacity to fit the 28K diverse facial samples, 
yielding collapsed reconstructions.

\begin{wraptable}{r}{0.55\textwidth}
    \centering
    \vspace{-12pt}
    \caption{%
        \textbf{Quantitative comparison on 2D INR reconstruction}
        on CelebA-HQ, ERA5, and ImageNet.
    }
    \label{tab:hyperinr-2d}
    \footnotesize
    \setlength{\tabcolsep}{4pt}
    \begin{tabular}{llcccc}
        \toprule
        Dataset & Metric & VAMoH & HyPoGen & LDMI & \textbf{Ours} \\
        \midrule
        \multirow{3}{*}{\textcolor{datasetgreen}{\textbf{CelebA-HQ}}}
            & PSNR $\uparrow$    & 21.3   & 16.2   & 24.8   & \textbf{27.7}   \\
            & SSIM $\uparrow$    & 0.6303 & 0.4550 & 0.6947 & \textbf{0.8635} \\
            & LPIPS $\downarrow$ & 0.1332 & 0.1463 & 0.0556 & \textbf{0.0296} \\
        \midrule
        \multirow{3}{*}{\textcolor{datasetgreen}{\textbf{ERA5}}}
            & PSNR $\uparrow$    & 39.0   & 40.0   & 44.6   & \textbf{49.3}   \\
            & SSIM $\uparrow$    & 0.8090 & 0.9947 & 0.9952 & \textbf{0.9991} \\
            & LPIPS $\downarrow$ & 0.1663 & 0.0008 & 0.0003 & \textbf{0.0001} \\
        \midrule
        \multirow{2}{*}{\textcolor{datasetgreen}{\textbf{ImageNet}}}
            & PSNR $\uparrow$ & 19.6   & 13.2   & 20.7   & \textbf{22.9}   \\
            & SSIM $\uparrow$ & 0.5955 & 0.2473 & 0.5610 & \textbf{0.6064} \\
        \bottomrule
    \end{tabular}
    \vspace{-8pt}
\end{wraptable}
\textbf{ERA5}~\citep{hersbach2019era5} provides hourly atmospheric measurements from ECMWF's climate reanalysis at 31\,km spatial resolution, with 8.5K training, 1.1K validation, and 2.4K test samples at $46{\times}90$ resolution; 
each pixel encodes a regional temperature. 
Our method again achieves the best score on all three metrics, confirming that the gains on natural images extend to scientific signals.

\begin{figure}[!t] 
\centering
\includegraphics[width=1.0\linewidth]{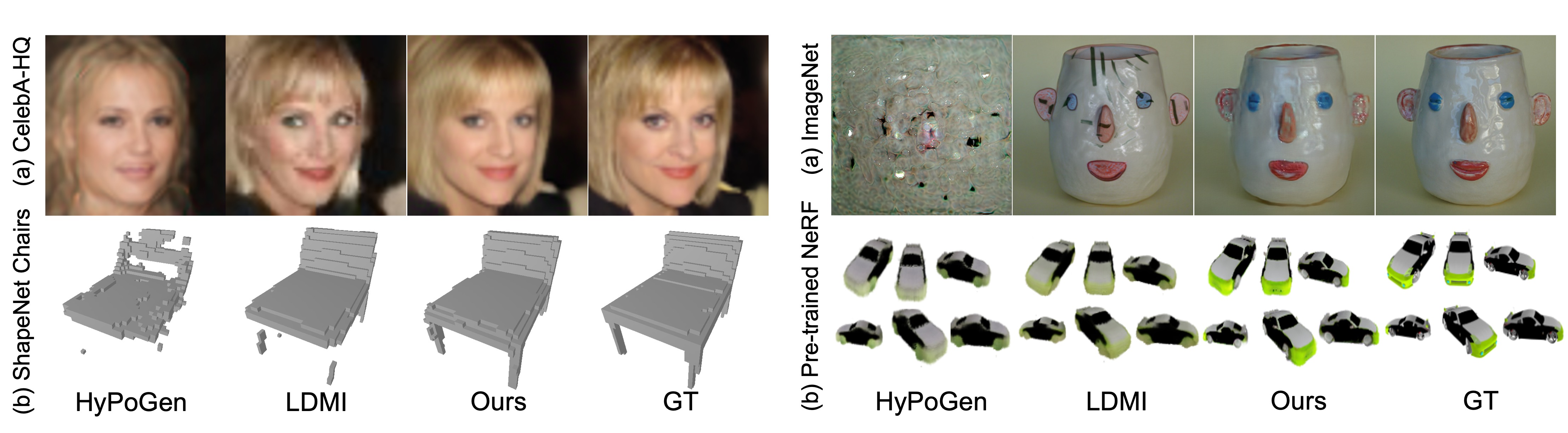}
\caption{Visual comparison of generated INRs across four datasets.
\textbf{Left:} 2D INR generation on (a) CelebA-HQ and 3D shape reconstruction on (b) ShapeNet Chairs. 
\textbf{Right:} 2D INR generation on (a) ImageNet and novel view synthesis on (b) Pre-trained NeRF. 
Columns show HyPoGen, LDMI, Ours, and the ground
truth (GT). 
Qualitative results on ERA5, as well as additional samples for all four datasets shown here, are provided in Appendix~\ref{app:additional-qualitative-results}.}
\vspace{-3pt}
\label{fig:2d_inrs}
\end{figure}

\textbf{ImageNet}~\citep{russakovsky2015imagenet} operates at a substantially larger scale. 
We use the ILSVRC 2012 subset, with 1.28M training, 50K validation, and 100K test images across 1{,}000 classes at $256{\times}256$ resolution following~\citep{peis2025hypertransforming}. 
Our method outperforms all baselines. HyPoGen fails to converge here, consistent with the scalability barrier of Sec.~\ref{sec:method:goal}. 
LDMI converges but produces visible artifacts and inconsistencies (Fig.~\ref{fig:2d_inrs}), 
as expected when per-layer weights are regressed without the chain-rule structure that the optimization-oriented principle provides.

\vspace{-2pt}
\subsection{3D INR Generation}
\label{sec:3d_eval}
\vspace{-1pt}

We consider two 3D INR generation tasks: 
geometric reconstruction on occupancy grids using ShapeNet Chairs~\citep{Chang2015ShapeNetAI},
and novel view synthesis on neural fields using the Pre-trained NeRF dataset~\citep{Ramirez2023DeepLO}, evaluated by occupancy grid accuracy and by PSNR with standard image-level metrics.

\vspace{-1pt}
\textbf{ShapeNet Chairs}~\citep{Chang2015ShapeNetAI} contains 6,778 chair instances (5.4K train, 677 val, 679 test).
Following~\citep{peis2025hypertransforming}, we adopt $32{\times}32{\times}32$ occupancy grids. Numerical results in Table~\ref{tab:hyperinr-3d} report occupancy accuracy and PSNR on rendered views, where our method achieves state-of-the-art performance on both metrics. 
Qualitatively (Fig.~\ref{fig:2d_inrs}), HyPoGen fails to capture fine details such as chair legs, LDMI yields incomplete reconstructions with coarse surfaces, and our method produces complete shapes with smooth and plausible geometry.

\vspace{-1pt}
\textbf{Pre-trained NeRF}~\citep{Ramirez2023DeepLO} comprises 103K objects (98.6K train, 4K test). 
Each object is equipped with 36 rendered views, camera parameters, and an individual NeRF pre-trained via ray-based optimization on its own images. 
For each task, we take 6 conditional views as input to synthesize 6 novel views. 
We evaluate three supervision configurations (Table~\ref{tab:hyperinr-3d}): 
\emph{Weights only} uses only the per-object NeRF weights from~\citep{Ramirez2023DeepLO}, \emph{Images only} uses only multi-view images, 
and \emph{Weights\,\&\,Images} uses both. 
\begin{wraptable}{r}{0.6\textwidth}
    \centering
    \vspace{-8pt}
    \caption{%
        \textbf{Quantitative comparison on 3D INR generation.}
        3D shape reconstruction on ShapeNet Chairs (top) and novel view synthesis on Pre-trained NeRF (bottom) under three
        supervision settings. ``HyperDiff.'' denotes HyperDiffusion.
    }
    \label{tab:hyperinr-3d}
    \footnotesize
    \setlength{\tabcolsep}{4pt}
    \begin{tabular}{llcccc}
        \toprule
        Dataset & Metric & VAMoH & HyPoGen & LDMI & \textbf{Ours} \\
        \midrule
        \multirow{2}{*}{\textcolor{datasetgreen}{\textbf{\shortstack[l]{ShapeNet\\Chairs}}}}
            & Accuracy $\uparrow$ & 96.8\% & 95.2\% & 97.3\% & \textbf{97.9\%} \\
            & PSNR $\uparrow$     & 15.6   & 19.0   & 22.4   & \textbf{23.8}   \\
        \specialrule{1pt}{2pt}{2pt}
        \textcolor{datasetgreen}{\textbf{\shortstack[l]{Pre-trained\\NeRF}}} & Metric & HyperDiff. & HyPoGen & LDMI & \textbf{Ours} \\
        \midrule
        Weights only & PSNR $\uparrow$ & 20.0 & 19.9 & 19.3 & \textbf{22.0} \\
        Images only  & PSNR $\uparrow$ & N/A  & 24.9 & 26.2 & \textbf{28.8} \\
        \midrule
        \multirow{3}{*}{\shortstack[l]{Weights\\\& Images}}
            & PSNR $\uparrow$    & 17.5   & 25.5   & 26.5   & \textbf{27.8}   \\
            & SSIM $\uparrow$    & 0.8774 & 0.921  & 0.929  & \textbf{0.948}  \\
            & LPIPS $\downarrow$ & 0.1972 & 0.1182 & 0.1079 & \textbf{0.0742} \\
        \bottomrule
    \end{tabular}
    \vspace{-8pt}
\end{wraptable}
HyperDiffusion is inapplicable to the images-only setting since it
diffuses directly in weight space (see Appendix~\ref{subsubsec:baseline methods}
for details).
Our method outperforms all baselines across different configurations. 
Our images-only result of 28.8\,dB is the highest overall in the table, showing that the framework does not rely on weight supervision -- a practical advantage given the cost of high-quality ground-truth INR weights. 
The gain over LDMI is largest under weights-only (+2.7\,dB), and the relative ranking of the two main baselines reverses there: 
HyPoGen (19.9\,dB) outperforms LDMI (19.3\,dB), contrary to their ordering elsewhere.
This matches the design regimes of Sec.~\ref{sec:method:goal}: when supervision is imposed directly in weight space,
optimization-oriented designs fit the target distribution better than attention-based hypernetworks without such structural inductive bias.

\vspace{-2pt}
\subsection{Computational Efficiency}
\label{sec:efficiency}
\vspace{-1pt}

Table~\ref{tab:batch16} reports parameter count, GPU memory, and inference latency at batch size 16 against HyPoGen and LDMI. 
Our hypernetwork uses fewer or comparable parameters across all five datasets, 
with a similar trend in GPU memory; the gap is largest on Pre-trained NeRF, where the baselines require 5--6$\times$ the memory. 
Inference latency is also lowest on the two largest benchmarks, 
ImageNet and Pre-trained NeRF. 
The accuracy gains in Tables~\ref{tab:hyperinr-2d} and~\ref{tab:hyperinr-3d} are therefore obtained without extra computational overhead, and at reduced cost on the most challenging tasks.

\begin{table}[h]
\centering
\footnotesize
\vspace{-12pt}
\caption{Computational efficiency evaluated at batchsize 16 on each dataset.}
\begin{tabular}{l ccc ccc ccc}
\toprule
& \multicolumn{3}{c}{\textbf{Params (M)}} & \multicolumn{3}{c}{\textbf{GPU Memory (GB)}} & \multicolumn{3}{c}{\textbf{Infer Time @ b=16 (ms)}} \\
\cmidrule(lr){2-4} \cmidrule(lr){5-7} \cmidrule(lr){8-10} 
\textbf{Dataset} & HyPoGen & LDMI & Ours & HyPoGen & LDMI & Ours & HyPoGen & LDMI & Ours \\
\midrule
CelebA-HQ      & 128.8 & 40.5 & 41.2 & 0.72 & 0.35 & 0.42 & 6.6   & 17.2  & 24.5 \\
ImageNet        & 578.7 & 544.9 & 124.5 & 8.20  & 8.12  & 5.51  & 66.8  & 84.5  & 58.5 \\
ERA5            & 93.1 & 48.9 & 21.3 & 0.59 & 0.51 & 0.33 & 21.2  & 25.0  & 24.6 \\
Shapes3D       & 8.3 & 30.9 & 4.1 & 0.82 & 0.79 & 0.80 & 4.7   & 8.3   & 10.2 \\
Pretrained NeRF & 1049.9 & 515.5 & 482.5 & 14.51 & 12.47 & 2.51 & 2382  & 2541  & 2018 \\
\bottomrule
\end{tabular}
\label{tab:batch16}
\vspace{-10pt}
\end{table}

\vspace{-1pt}
\subsection{Ablation and Analysis}
\label{sec:analysis}
\vspace{-1pt}

We ablate the two architectural principles 
of Sec.~\ref{sec:method} and validate the scalability claim of Sec.~\ref{sec:method:goal} empirically.
Ablations on the forward chain, the backward decomposition, and the number of refinement steps $T$ are in Appendix~\ref{app:more-ablations}.

\begin{table}[!htbp]
\centering
\footnotesize
\setlength{\tabcolsep}{5pt}
\caption{Ablations on the two architectural principles of Sec.~\ref{sec:method}. 
PSNR ($\uparrow$) on CelebA-HQ, ImageNet, and Pre-trained NeRF (\emph{Weights\,\&\,Images}).}
\label{tab:design-ablation}
\begin{tabular}{l|ccc}
\toprule
\textbf{Variant} & CelebA-HQ & ImageNet & NeRF \\
\midrule
\textbf{Ours (Full)}                  & \textbf{27.7} & \textbf{22.9} & \textbf{27.8} \\
\midrule
Column-wise Tokenization              & 25.3 & 19.8 & 25.1 \\
Reversed Q-KV                         & 23.5 & 18.3 & 22.9 \\
\bottomrule
\end{tabular}
\end{table}
\paragraph{Output-neuron-centric tokenization.}
\textit{Column-wise Tokenization} replaces the
output-neuron-centric scheme of Sec.~\ref{sec:method:tokenization} with the input-neuron-centric scheme of Fig.~\ref{fig:tokenization}: 
each token of
$\boldsymbol{\omega}_i$ encodes a column of $W_i$ rather than a row, 
breaking the correspondence between the $j$-th token and the $j$-th output neuron of layer $i$, while every other component is kept unchanged. 
PSNR drops by 2.4 / 3.1 / 2.7\,dB on
CelebA-HQ / ImageNet / NeRF
(Table~\ref{tab:design-ablation}), 
confirming that grouping the parameters that produce one output neuron into a co-adapted token,
rather than the parameters that fan out from one input neuron to unrelated output neurons, 
is what aligns the token axis with the output-neuron axis on which all five per-layer quantities of Sec.~\ref{sec:method:derivation} live.

\vspace{-1pt}
\paragraph{Information-flow consistency in QKV assignment.}
\textit{Reversed Q-KV} 
swaps $Q$ with $K$/$V$ in every cross-attention module of the iterative module, violating the information-flow-consistency rule of Sec.~\ref{sec:method:iterative-module}. 
The forward block, 
for instance, becomes $\boldsymbol{\tau}_i = \mathrm{CrossAttn}(
\boldsymbol{\tau}_{i-1}, \boldsymbol{\omega}_i,
\boldsymbol{\omega}_i)$, which places $V$ in the parameter space and so writes its output into a role inconsistent with the pseudo-activation position of $\boldsymbol{\tau}_i$; 
the four backward modules
(Eqs.~\ref{eq:method-Jtau}--\ref{eq:method-delta-grad}) are reversed analogously. 
PSNR drops by 4.2 / 4.6 / 4.9\,dB across the three datasets (Table~\ref{tab:design-ablation}). 
Since only the QKV assignment changes and all parameter counts are preserved, the gap directly measures the value of the principle.

\begin{figure}[!htbp]
\centering
\begin{subfigure}[t]{0.49\linewidth}
\centering
\includegraphics[width=\linewidth]{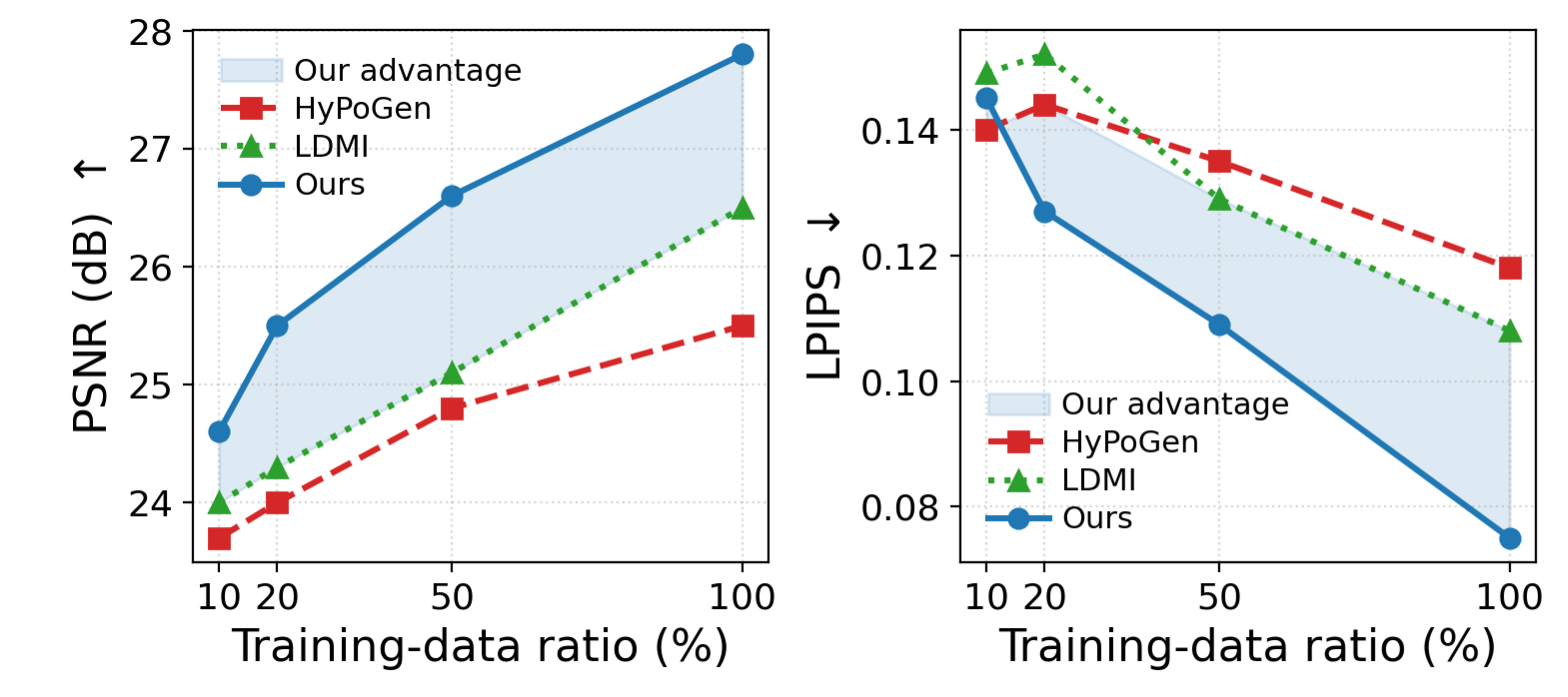}
\caption{Effect of training-data scale.}
\label{fig:scalability_comparison}
\end{subfigure}
\hfill
\begin{subfigure}[t]{0.49\linewidth}
\centering
\includegraphics[width=\linewidth]{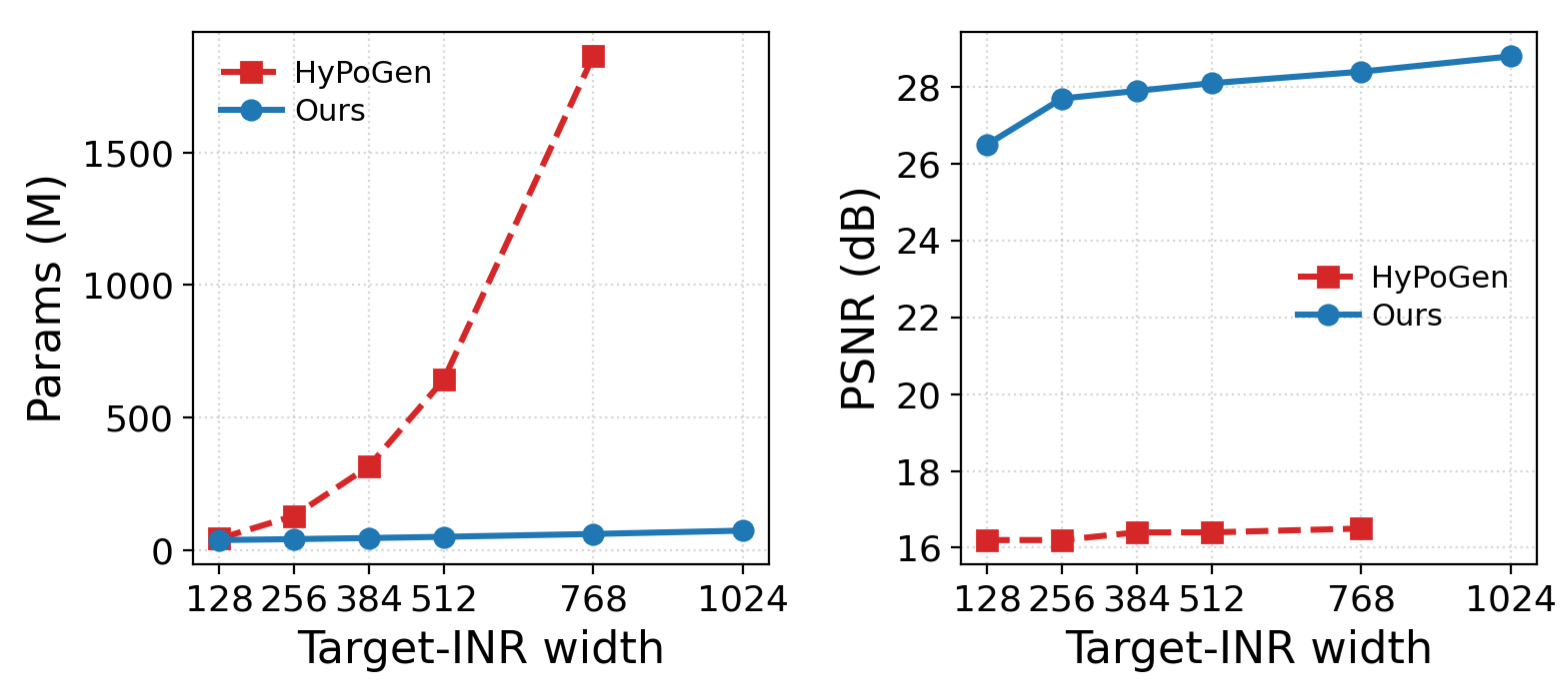}
\caption{Effect of target-INR width.}
\label{fig:width-scaling}
\end{subfigure}
\vspace{-2pt}
\caption{\textbf{Empirical scalability analysis.}
(a) Performance versus training-data ratio on NeRF generation, testing how accuracy improves with data scale.
(b) Parameter count and PSNR versus target-INR width on CelebA-HQ at batch size 64, testing the parameter-count prediction of Eq.~\ref{eq:app-contrast-compressed}: HyPoGen's parameter count
grows rapidly while ours grows mildly with much more gains in PSNR.}
\label{fig:scalability}
\vspace{-5pt}
\end{figure}

\vspace{-1pt}
\paragraph{Scalability Analysis}
\label{sec:scalability}
The parameter-count derivation in Appendix~\ref{app:param-count} formalizes the scalability barrier mentioned in Sec.~\ref{sec:method:goal}: per target-network layer per refinement step, HyPoGen's core body costs
$\mathcal{O}(n_i^2 \cdot d^2)$ parameters while ours costs $\mathcal{O}(d^2)$, the factor of $n_i^2$ being the structural difference (Eq.~\ref{eq:app-contrast-compressed}). 
We test the two consequences empirically: \emph{data scaling}
(Fig.~\ref{fig:scalability_comparison}) 
and \emph{width scaling}
(Fig.~\ref{fig:width-scaling}).

\vspace{-1pt}
\paragraph{Data scaling.}
Fig.~\ref{fig:scalability_comparison} shows how each method's performance evolves as the fraction of Pre-trained NeRF training data grows from 10\% to 100\%. 
All three methods improve, 
but at very different rates: 
our method gains $+3.2$\,dB PSNR over the range, LDMI $+2.5$\,dB, and HyPoGen only $+1.8$\,dB; 
LPIPS follows the same ranking. 
The gap between our method and LDMI also grows with scale, from $+0.6$\,dB at 10\% to $+1.3$\,dB at full data,
indicating that the methods diverge precisely in the regime that matters for amortized INR generation. 
HyPoGen saturates earliest, consistent with the $\mathcal{O}\!\left(n_i^2 \cdot d^2\right)$ factor of Eq.~\ref{eq:app-hypogen-compressed} bounding how much new structure its MLP body can absorb. 
LDMI scales further but is bounded above by ours: an attention body alone is not sufficient,
since without the optimization-oriented structure of Sec.~\ref{sec:method:iterative-module}, additional data refines each layer's weights in isolation rather than the joint forward-backward computation that produces a coherent target INR.

\vspace{-1pt}
\paragraph{Width scaling.}
Fig.~\ref{fig:width-scaling} tests the parameter-count prediction of Eq.~\ref{eq:app-contrast-compressed} empirically.
The prediction is that HyPoGen's core body carries a factor of $n_i^2$ and ours does not; 
the test sweeps the target-INR width from 128 to 1024 on CelebA-HQ at batch size 64, comparing HyPoGen and ours -- the two methods that both encode the optimization-oriented principle, 
isolating the MLP body versus the attention body as the architectural variable. 
The prediction is borne out: HyPoGen's parameter count grows rapidly with width and exhausts GPU memory at 1024, 
while ours grows mildly across the entire range. 
PSNR follows the same pattern: ours improves from
26.5 to 28.8\,dB as width grows, while HyPoGen remains around 16\,dB across all tractable widths and fails outright at 1024.

\vspace{-3pt}
\section{Conclusion}
\label{sec:conclusion}
\vspace{-2pt}
We have shown that 
the scalability barrier of 
optimization-oriented hypernetworks 
for INR generation is a property 
of the MLP body
rather than of the principle 
of preserving chain-rule structure,
and that replacing the body with 
attention dissolves the barrier
without compromising the principle. 
The broader observation, 
for which our empirical results provide direct evidence, is that
single-pass scalability and chain-rule structure are not opposing design goals; 
they are simultaneously achievable through 
a combination of weight-matrix tokenization, 
role-driven module specification, 
and a token-space rollout that mirrors one optimization step. 
The same construction is likely to apply
wherever a learned model amortizes a structured optimization procedure, including meta-learning, learned optimizers, and structured weight-generation tasks beyond INRs.

\clearpage
\section*{Acknowledgments}
We thank the anonymous reviewers for their valuable comments and suggestions. This work was supported by the Early Career Scheme of the Research Grants Council (grant no. 27207224), the HKU-100 Award, and a donation from the Musketeers Foundation. This work was also supported in part by the JC STEM Lab of Autonomous Intelligent Systems funded by The Hong Kong Jockey Club Charities Trust. We acknowledge the computing resources provided by the HKU-Shanghai Intelligent Computing Research Center.

\bibliographystyle{unsrt}
\bibliography{neurips_2026}

\newpage

\section{Appendix}

\subsection{Parameter-Count Derivation for the Core Iterative Body}
\label{app:param-count}

This section expands the parameter-count comparison summarized in Sec.~\ref{sec:method:goal} and Sec.~\ref{sec:method:rollout}.
We derive the per-layer per-refinement-step parameter count of HyPoGen's MLP-based core and of our attention-based core, in two formulations: in the original target-network dimensions $n_i$ and $n_{i-1}$, 
and under the row-level compression to embedding dimension $d$ that the parameter tokens of Sec.~\ref{sec:method:tokenization} implement. 
Throughout, we count parameters of the \emph{core iterative body} only, that is, the sub-modules that consume a representation of $W_i$ and produce an update of the same shape. 
The per-layer encoders and decoders that convert between $W_i$ and the token representation are shared between the two architectures and excluded from the count.

\subsubsection{Original target-network dimensions}
\label{app:param-count:original}

Consider a single fully connected layer of an MLP target network with weight matrix $W_i \in \mathbb{R}^{n_i \times n_{i-1}}$. 
Without any compression, the layer is presented to the hypernetwork's core body as a flat vector of $n_i \cdot n_{i-1}$ scalars, and the body must produce an update of the same dimension.

\paragraph{HyPoGen's MLP body.}
HyPoGen's core body is a stack of linear transformations that operate on this flat representation. 
Each linear transformation maps a vector of dimension $n_i \cdot n_{i-1}$ to a vector of the same dimension and therefore carries $(n_i \cdot n_{i-1})^2 = n_i^2 \cdot n_{i-1}^2$ parameters. 
The parameter count per layer per refinement step is therefore
\begin{equation}
\label{eq:app-hypogen-original}
\mathcal{C}^{\mathrm{HyPoGen}}_{\mathrm{orig}}
\;=\; \mathcal{O}\!\left(n_i^2 \cdot n_{i-1}^2\right),
\end{equation}
quartic in target-network width and growing with both layer dimensions $n_i$ and $n_{i-1}$.

\paragraph{Our attention body.}
Our core body represents $W_i$ as a token sequence of length $n_i$ in which each token is the $n_{i-1}$-dimensional encoding
of one row of $W_i$, following the output-neuron-centric tokenization of Sec.~\ref{sec:method:tokenization}. 
Five cross-attention modules operate on token sequences of this shape: one forward module $\phi_i$ and four backward modules
$\psi^{\boldsymbol{\tau}}_i, \psi^{\boldsymbol{\omega}}_i,
\psi^{\boldsymbol{\delta}}_i, \psi^{\nabla}_i$. 
Each module has
query, key, value, and output projections of size $n_{i-1} \times n_{i-1}$, contributing $\mathcal{O}(n_{i-1}^2)$ parameters in total, and the same projections are applied across all $n_i$ tokens by the attention mechanism rather than parameterized separately for each token. 
The number $n_i$ therefore enters as the length of the sequence the same module is applied to, not as a factor in the module's own parameter count. 
With a constant number of modules per refinement step,
the parameter count per layer per refinement step is
\begin{equation}
\label{eq:app-ours-original}
\mathcal{C}^{\mathrm{Ours}}_{\mathrm{orig}}
\;=\; \mathcal{O}\!\left(n_{i-1}^2\right),
\end{equation}
independent of $n_i$ and quadratic rather than quartic in $n_{i-1}$.

\paragraph{Headline contrast.}
Combining
Eqs.~\ref{eq:app-hypogen-original}--\ref{eq:app-ours-original},
the contrast in the original dimensions is
\begin{equation}
\label{eq:app-contrast-original}
\underbrace{\mathcal{O}(n_i^2 \cdot n_{i-1}^2)}_{\text{HyPoGen}}
\quad\longrightarrow\quad
\underbrace{\mathcal{O}(n_{i-1}^2)}_{\text{Ours}}.
\end{equation}
The structural difference is the factor of $n_i^2$, which our hypernetwork removes because $n_i$ governs the length of the token sequence processed by a module whose parameter count is independent of that length, 
rather than entering as a multiplier on the dimension of a flat vector whose representation requires a parameter count that grows with $n_i$.

\subsubsection{Compression to embedding dimension $d$}
\label{app:param-count:compressed}

The cost of either core can be reduced further by compressing each row of $W_i$ from $n_{i-1}$ scalars to a $d$-dimensional
latent through the per-layer encoder, with the per-layer decoder recovering the row when the actual weights are needed. 
Under this compression, 
the layer is presented to the core body as either a flat vector of dimension $n_i \cdot d$ (for HyPoGen) or a token sequence of length $n_i$ with each token of dimension $d$ (for ours).

\paragraph{HyPoGen's MLP body under compression.}
A linear transformation between two spaces of dimension $n_i \cdot d$ carries $(n_i \cdot d)^2 = n_i^2 \cdot d^2$ parameters,
giving a parameter count per layer per refinement step of
\begin{equation}
\label{eq:app-hypogen-compressed}
\mathcal{C}^{\mathrm{HyPoGen}}_{\mathrm{cmp}}
\;=\; \mathcal{O}\!\left(n_i^2 \cdot d^2\right).
\end{equation}
The factor of $d^2$ replaces the factor of $n_{i-1}^2$ from Eq.~\ref{eq:app-hypogen-original}, and the factor of $n_i^2$ remains because the MLP body still processes $n_i$ rows through a transformation whose parameter count grows with $n_i$.

\paragraph{Our attention body under compression.}
Each of the five cross-attention modules now has projections of size $d \times d$, contributing $\mathcal{O}(d^2)$ parameters per module, and the same projections are applied across all $n_i$ tokens. 
The parameter count per layer per refinement step is
\begin{equation}
\label{eq:app-ours-compressed}
\mathcal{C}^{\mathrm{Ours}}_{\mathrm{cmp}}
\;=\; \mathcal{O}\!\left(d^2\right),
\end{equation}
independent of both $n_i$ and $n_{i-1}$. 
The dependence on target-network width is removed entirely; the embedding dimension $d$ is a hypernetwork hyperparameter chosen independently of the target-network width and controls hypernetwork capacity in isolation.

\paragraph{Headline contrast under compression.}
The contrast under compression is
\begin{equation}
\label{eq:app-contrast-compressed}
\underbrace{\mathcal{O}(n_i^2 \cdot d^2)}_{\text{HyPoGen}}
\quad\longrightarrow\quad
\underbrace{\mathcal{O}(d^2)}_{\text{Ours}}.
\end{equation}
The factor of $n_i^2$ is again the structural difference between the two architectures, and the mechanism that removes it is the same as in the uncompressed case: 
$n_i$ governs sequence length rather than module parameter count.

\subsubsection{Summary}
\label{app:param-count:summary}

Table~\ref{tab:app-param-count} summarizes the four counts derived above. 
In both formulations, the parameter count of HyPoGen's core body retains a factor of $n_i^2$ that our attention-based core does not, 
and this factor is the structural origin of the scalability barrier identified in Sec.~\ref{sec:method:goal}. 
The empirical confirmation of these scalings -- parameter count, GPU memory, and PSNR as the target-network width $N$ varies from $128$ to $1024$ -- is
reported in Sec.~\ref{sec:scalability}
(Fig.~\ref{fig:width-scaling}).

\begin{table}[h]
\caption{Parameter count of the core iterative body per target-network layer per refinement step, for HyPoGen's MLP-based core and our attention-based core, in the original target-network dimensions and under row-level compression to embedding dimension $d$. 
Encoders and decoders are shared between the two architectures and excluded from the count.}
\centering
\small
\begin{tabular}{lll}
\toprule
& Original dimensions & Under compression \\
\midrule
HyPoGen (MLP body)
& $\mathcal{O}(n_i^2 \cdot n_{i-1}^2)$
& $\mathcal{O}(n_i^2 \cdot d^2)$ \\
Ours (attention body)
& $\mathcal{O}(n_{i-1}^2)$
& $\mathcal{O}(d^2)$ \\
\midrule
Structural difference
& factor of $n_i^2$
& factor of $n_i^2$ \\
\bottomrule
\end{tabular}
\label{tab:app-param-count}
\end{table}

\subsection{Forward-Block Correspondence with a Standard INR Layer}
\label{app:forward-correspondence}

This section expands the verification claim in Sec.~\ref{sec:method:iterative-module} that the cross-attention module $\phi_i$ derived from information-flow consistency reproduces the structural skeleton of a standard INR layer (Eq.~\ref{eq:method-fwd}) component by component. 
We work through each component in turn and summarize the correspondence in Table~\ref{tab:app-forward-correspondence}.

\paragraph{Linear mixing.}
A standard INR layer mixes the previous activation $\mathbf{h}_{i-1} \in \mathbb{R}^{n_{i-1}}$ through the rows of $W_i \in \mathbb{R}^{n_i \times n_{i-1}}$ to produce the pre-activation $W_i \mathbf{h}_{i-1} \in \mathbb{R}^{n_i}$,
whose $j$-th component is $W_i^{(j,:)} \mathbf{h}_{i-1}$, 
the inner product of the $j$-th row of $W_i$ with the entire previous activation. 
The cross-attention module $\phi_i$ in Eq.~\ref{eq:method-forward} produces a score matrix $\boldsymbol{\omega}_i \boldsymbol{\tau}_{i-1}^\top / \sqrt{d}$,
whose $(j, k)$-entry is the inner product $\boldsymbol{\omega}_{i,j}^\top \boldsymbol{\tau}_{i-1, k} / \sqrt{d}$ between the $j$-th parameter token and the $k$-th previous-layer pseudo-activation token. 
Both operations therefore form one inner product per (output-neuron, input-neuron) pair, 
between the parameters that produce the output neuron and the representation of the input neuron's activation. 
The factor of $1 / \sqrt{d}$ is a normalization specific to attention that has no counterpart in a standard layer; 
it does not affect the structural correspondence.

\paragraph{Non-linearity.}
A standard INR layer applies the activation function $\sigma$ element-wise to $W_i \mathbf{h}_{i-1}$, producing $\mathbf{h}_i = \sigma(W_i \mathbf{h}_{i-1})$. 
The cross-attention module $\phi_i$ applies a row-wise softmax to the score matrix.
Both operations introduce a non-linearity between the linear mixing and the next stage of the computation; 
the softmax differs from $\sigma$ in its normalization (it produces a convex
combination, $\sigma$ does not) 
and in its row-wise rather than element-wise structure, but both serve the structural role of a non-linearity in the layer.

\paragraph{Output-unit alignment.}
A standard INR layer satisfies an output-unit alignment property: 
the $j$-th component of $\mathbf{h}_i$ is produced by the $j$-th row of $W_i$ alone, and consequently the $j$-th output neuron inherits its identity from the $j$-th row. 
The cross-attention module $\phi_i$ inherits the same property from the row-wise tokenization of Sec.~\ref{sec:method:tokenization}: 
the $j$-th output token of $\phi_i$ is produced by the $j$-th row of the score matrix and, 
after softmax, 
by the row-wise convex combination of $\boldsymbol{\tau}_{i-1}$ that the $j$-th query token $\boldsymbol{\omega}_{i,j}$ specifies. 
The $j$-th query token is the encoding of the $j$-th row of $W_i$, so the $j$-th output token of $\phi_i$ is in one-to-one correspondence with the $j$-th output neuron of the target layer.

\begin{table}[h]
\caption{Component-by-component correspondence between a standard
INR layer (Eq.~\ref{eq:method-fwd}) and the cross-attention module $\phi_i$ (Eq.~\ref{eq:method-forward}) that realizes the forward block.}
\centering
\small
\begin{tabular}{lll}
\toprule
& Standard layer & Cross-attention module $\phi_i$ \\
\midrule
Linear mixing
& $W_i \, \mathbf{h}_{i-1}$
& $\boldsymbol{\omega}_i \, \boldsymbol{\tau}_{i-1}^\top /
\sqrt{d}$ \\
Non-linearity
& activation $\sigma$
& row-wise softmax \\
Output-unit alignment
& $j$-th row $\to$ $j$-th neuron
& $j$-th query $\to$ $j$-th output token \\
\bottomrule
\end{tabular}
\label{tab:app-forward-correspondence}
\end{table}

\subsection{Per-Module QKV Derivations for the Backward Block}
\label{app:qkv-derivations}

This section justifies the four output-representation claims that Sec.~\ref{sec:method:iterative-module} uses to fix the QKV assignments of the backward modules
$\psi^{\boldsymbol{\tau}}_i, \psi^{\boldsymbol{\omega}}_i,
\psi^{\boldsymbol{\delta}}_{i-1}, \psi^{\nabla}_i$ 
by information-flow consistency. 
For each module, we identify the target-network quantity it surrogates, 
identify the representation that quantity's output lives in, and read off the QKV assignment.

\paragraph{Activation-Jacobian module $\psi^{\boldsymbol{\tau}}_i$.}
The module surrogates the activation Jacobian $\partial \mathbf{h}_i / \partial \mathbf{h}_{i-1}$, 
which appears in Eq.~\ref{eq:method-chain} as the linear map that transforms the layer-$i$ loss-signal $\delta_i$ into the layer-$(i-1)$ loss-signal $\delta_{i-1} = \delta_i \cdot \partial \mathbf{h}_i / \partial \mathbf{h}_{i-1}$. 
The output of this contraction is in the layer-$(i-1)$ activation representation,
since it is the gradient of $\mathcal{L}$ with respect to $\mathbf{h}_{i-1}$. 
The activation-Jacobian token sequence
$\boldsymbol{J}^{\boldsymbol{\tau}}_{i-1}$ that $\psi^{\boldsymbol{\tau}}_i$ produces therefore lives in this representation. 
Among the module's two inputs $\boldsymbol{\omega}_i$ and $\boldsymbol{\tau}_{i-1}$, 
the sequence $\boldsymbol{\tau}_{i-1}$ is the layer-$(i-1)$ pseudo-activation token sequence and already lives in the required representation, 
while $\boldsymbol{\omega}_i$ lives in the layer-$i$ parameter representation. 
By information-flow consistency, $V = K = \boldsymbol{\tau}_{i-1}$ and $Q = \boldsymbol{\omega}_i$.

\paragraph{Parameter-Jacobian module $\psi^{\boldsymbol{\omega}}_i$.}
The module surrogates the parameter Jacobian $\partial \mathbf{h}_i / \partial W_i$, 
which appears in Eq.~\ref{eq:method-chain} as the linear map that transforms $\delta_i$ into the parameter gradient $\nabla_{W_i}\mathcal{L} = \delta_i \cdot \partial \mathbf{h}_i / \partial W_i$. 
The output of this contraction is in the layer-$i$ parameter representation, 
since it is the gradient of $\mathcal{L}$ with respect to $W_i$. 
The parameter-Jacobian token sequence $\boldsymbol{J}^{\boldsymbol{\omega}}_i$ that $\psi^{\boldsymbol{\omega}}_i$ produces therefore lives in this representation. 
The module's two inputs are again
$\boldsymbol{\omega}_i$ and $\boldsymbol{\tau}_{i-1}$, 
but now
$\boldsymbol{\omega}_i$ is the input that lives in the layer-$i$ parameter representation. 
By information-flow consistency, 
$V = K = \boldsymbol{\omega}_i$ and $Q = \boldsymbol{\tau}_{i-1}$, 
the QKV roles are reversed relative to
$\psi^{\boldsymbol{\tau}}_i$ even though the two inputs are the same.

\paragraph{Loss-signal-propagation module
$\psi^{\boldsymbol{\delta}}_{i-1}$.}
The module realizes the chain-rule combination $\delta_{i-1} =
\delta_i \cdot \partial \mathbf{h}_i / \partial \mathbf{h}_{i-1}$
of Eq.~\ref{eq:method-chain}, which produces the layer-$(i-1)$
loss-signal. The output token sequence
$\boldsymbol{\delta}_{i-1}$ therefore lives in the layer-$(i-1)$
activation representation. The module's two inputs are
$\boldsymbol{\delta}_i$ and $\boldsymbol{J}^{\boldsymbol{\tau}}_{i-1}$.
The sequence $\boldsymbol{\delta}_i$ lives in the layer-$i$
activation representation; the sequence
$\boldsymbol{J}^{\boldsymbol{\tau}}_{i-1}$ was just established
to live in the layer-$(i-1)$ activation representation. By
information-flow consistency, $V = K =
\boldsymbol{J}^{\boldsymbol{\tau}}_{i-1}$ and $Q =
\boldsymbol{\delta}_i$.

\paragraph{Parameter-gradient module $\psi^{\nabla}_i$.}
The module realizes the chain-rule combination
$\nabla_{W_i}\mathcal{L} = \delta_i \cdot \partial \mathbf{h}_i /
\partial W_i$ of Eq.~\ref{eq:method-chain}, which produces the
parameter gradient. The output token sequence
$\nabla_{\boldsymbol{\omega}_i}\mathcal{L}$ therefore lives in
the layer-$i$ parameter representation. The module's two inputs
are $\boldsymbol{\delta}_i$ and
$\boldsymbol{J}^{\boldsymbol{\omega}}_i$. The sequence
$\boldsymbol{\delta}_i$ lives in the layer-$i$ activation
representation; the sequence
$\boldsymbol{J}^{\boldsymbol{\omega}}_i$ was just established to
live in the layer-$i$ parameter representation. By
information-flow consistency, $V = K =
\boldsymbol{J}^{\boldsymbol{\omega}}_i$ and $Q =
\boldsymbol{\delta}_i$.

\paragraph{Summary.}
The four QKV assignments derived above are collected in
Table~\ref{tab:app-qkv-derivations} and match the color coding in
Fig.~\ref{fig:iterative_module}. Each assignment follows from
information-flow consistency once the output representation is
identified from the target-network quantity the module
surrogates; no module-specific design choice enters beyond the
identification of the output representation.

\begin{table}[h]
\caption{QKV assignments of the four backward modules, derived
from information-flow consistency given the output
representation each module's target-network counterpart lives
in.}
\centering
\small
\begin{tabular}{lllll}
\toprule
Module & Surrogate of & Output representation & $V = K$ & $Q$
\\
\midrule
$\psi^{\boldsymbol{\tau}}_i$
& $\partial \mathbf{h}_i / \partial \mathbf{h}_{i-1}$
& layer-$(i-1)$ activation
& $\boldsymbol{\tau}_{i-1}$
& $\boldsymbol{\omega}_i$ \\
$\psi^{\boldsymbol{\omega}}_i$
& $\partial \mathbf{h}_i / \partial W_i$
& layer-$i$ parameter
& $\boldsymbol{\omega}_i$
& $\boldsymbol{\tau}_{i-1}$ \\
$\psi^{\boldsymbol{\delta}}_{i-1}$
& $\delta_i \cdot \partial \mathbf{h}_i / \partial
\mathbf{h}_{i-1}$
& layer-$(i-1)$ activation
& $\boldsymbol{J}^{\boldsymbol{\tau}}_{i-1}$
& $\boldsymbol{\delta}_i$ \\
$\psi^{\nabla}_i$
& $\delta_i \cdot \partial \mathbf{h}_i / \partial W_i$
& layer-$i$ parameter
& $\boldsymbol{J}^{\boldsymbol{\omega}}_i$
& $\boldsymbol{\delta}_i$ \\
\bottomrule
\end{tabular}
\label{tab:app-qkv-derivations}
\end{table}

\subsection{Per-Layer Encoder and Decoder}
\label{app:method:encode-decode}

The tokenization scheme of Sec.~\ref{sec:method:tokenization} introduces a per-layer encoder $\mathrm{Encode}_i$ and a per-layer decoder $\mathrm{Decode}_i$ for layer $i$ of the target network. They sit at the two ends of the iterative module: $\mathrm{Encode}_i$ converts each row of $W_i$ into one parameter token to form $\boldsymbol{\omega}_i$, and $\mathrm{Decode}_i$ recovers each row at the end of the rollout (Eq.~\ref{eq:method-hypernetwork}). This appendix specifies both.

The encoder maps the $j$-th row of $W_i$ to the $j$-th parameter token through an MLP, adds a learnable positional embedding $\boldsymbol{p}_{i,j} \in \mathbb{R}^{d}$ to distinguish the row identity, and applies layer normalization:
\begin{equation}
\label{eq:app-encode}
\boldsymbol{\omega}_{i,j} \;=\; \mathrm{Encode}_i\!\big(W_i^{(j,:)}\big) \;=\; \mathrm{LayerNorm}\!\Big( \mathrm{MLP}^{\mathrm{enc}}_{i}\!\big(W_i^{(j,:)}\big) \;+\; \boldsymbol{p}_{i,j} \Big), \qquad j = 1, \dots, n_i.
\end{equation}
The decoder inverts this map row by row through layer normalization followed by an MLP:
\begin{equation}
\label{eq:app-decode}
W_i^{(j,:)} \;=\; \mathrm{Decode}_i(\boldsymbol{\omega}_{i,j}) \;=\; \mathrm{MLP}^{\mathrm{dec}}_{i}\!\big(\mathrm{LayerNorm}(\boldsymbol{\omega}_{i,j})\big), \qquad j = 1, \dots, n_i.
\end{equation}
Both $\mathrm{MLP}^{\mathrm{enc}}_i: \mathbb{R}^{n_{i-1}} \!\to\! \mathbb{R}^{d}$ and $\mathrm{MLP}^{\mathrm{dec}}_i: \mathbb{R}^{d} \!\to\! \mathbb{R}^{n_{i-1}}$ are shared across the $n_i$ rows of layer $i$, so each module's parameter count is independent of $n_i$; the rows are distinguished only by the positional embeddings $\boldsymbol{p}_{i,1}, \dots, \boldsymbol{p}_{i,n_i}$, which contribute a term that scales linearly in $n_i$ rather than quadratically.

\subsection{Initialization Sub-modules}
\label{app:init}

The full hypernetwork $\mathcal{H}_\Omega$ specified in Sec.~\ref{sec:method:rollout} requires three initial token sequences for the iterative module of Sec.~\ref{sec:method:iterative-module} to be unrolled: the pseudo-activation $\boldsymbol{\tau}_0$ that seeds the forward chain at the input side of every iteration, the loss-signal $\boldsymbol{\delta}_L^t$ that seeds the backward block at the output side of iteration $t$, and the parameter token sequence $\boldsymbol{\omega}_i^{0}$ at each layer $i$ that seeds the parameter-token rollout of Eq.~\ref{eq:method-refinement-step}. The three sub-modules differ in when they are evaluated: $\boldsymbol{\tau}_0$ depends only on $\mathbf{z}_\zeta$ and is computed once at the start of the rollout, $\boldsymbol{\delta}_L^t$ depends on the layer-$L$ pseudo-activation $\boldsymbol{\tau}_L^t$ and is therefore re-evaluated at every iteration $t$, and $\boldsymbol{\omega}_i^{0}$ is consumed only at $t=0$ and overwritten by the rollout's update rule at every subsequent iteration. This appendix specifies the three sub-modules that produce them. All three take the conditioning input $\mathbf{z}_\zeta$ as input and emit token sequences in the format expected by the iterative module.

\subsubsection{Pseudo-activation Initialization $\boldsymbol{\tau}_0$}
\label{app:init:tau}

The pseudo-activation tokens $\boldsymbol{\tau}_0 \in \mathbb{R}^{n_0 \times d}$ play the role at the input of the forward chain that the input coordinate $\mathbf{x}$ plays in the target INR's forward pass (Eq.~\ref{eq:method-fwd}). They are produced by a sequence of $n_0$ learnable query tokens $\boldsymbol{Q}^{\boldsymbol{\tau}} \in \mathbb{R}^{n_0 \times d}$ cross-attending to $\mathbf{z}_\zeta$:
\begin{equation}
\label{eq:app-init-tau}
\boldsymbol{\tau}_0 \;=\; \mathrm{CrossAttn}_{\boldsymbol{\tau}}\big(\boldsymbol{Q}^{\boldsymbol{\tau}}, \mathbf{z}_\zeta, \mathbf{z}_\zeta\big) \;\in\; \mathbb{R}^{n_0 \times d},
\end{equation}
shown as the leftmost CA block in Fig.~\ref{fig:iterative_module}. The query tokens supply the row count $n_0$ — the number of input neurons of the target INR — and absorb task-specific content from $\mathbf{z}_\zeta$ through the cross-attention's keys and values. The same $\boldsymbol{\tau}_0$ is reused as the input to $\phi_1$ at every iteration $t$, since the construction depends only on $\mathbf{z}_\zeta$.

\subsubsection{Loss-signal Initialization $\boldsymbol{\delta}_L^t$}
\label{app:init:delta}

The loss-signal tokens $\boldsymbol{\delta}_L^t \in \mathbb{R}^{n_L \times d}$ play the role at the output of the backward block that the output-layer loss-signal $\delta_L = \partial \mathcal{L} / \partial \mathbf{h}_L$ plays in the target INR's backward pass. At the target network, $\delta_L$ depends on the prediction $\mathbf{h}_L$ and on the supervision target through the reconstruction loss $\mathcal{L}$. The hypernetwork mirrors this dependence by feeding the $\boldsymbol{\delta}_L^t$ sub-module with two inputs that carry the same two information sources in token form: the pseudo-activation $\boldsymbol{\tau}_L^t$, which is the prediction-side analogue at iteration $t$, and the conditioning input $\mathbf{z}_\zeta$, which is the signal-side analogue and carries the supervision target through the upstream encoder that produces $\mathbf{z}_\zeta$ from $\zeta$. The two inputs are concatenated along the token axis, and a sequence of $n_L$ learnable query tokens $\boldsymbol{Q}^{\boldsymbol{\delta}} \in \mathbb{R}^{n_L \times d}$ cross-attends to the concatenated sequence:
\begin{equation}
\label{eq:app-init-delta}
\boldsymbol{\delta}_L^t \;=\; \mathrm{CrossAttn}_{\boldsymbol{\delta}}\big(\boldsymbol{Q}^{\boldsymbol{\delta}}, \boldsymbol{C}^t, \boldsymbol{C}^t\big) \;\in\; \mathbb{R}^{n_L \times d},
\qquad
\boldsymbol{C}^t = \big[\,\boldsymbol{\tau}_L^t \;;\; \mathbf{z}_\zeta\,\big],
\end{equation}
shown as the concat-plus-CA block on the right of Fig.~\ref{fig:iterative_module}. The query tokens supply the row count $n_L$ to match $\boldsymbol{\tau}_L^t$ token-by-token, and absorb both prediction-side and signal-side content from the concatenated keys and values. Unlike $\boldsymbol{\tau}_0$, this sub-module is re-evaluated at every iteration $t$ because $\boldsymbol{\tau}_L^t$ changes between iterations as the parameter tokens are refined.

\subsubsection{Parameter Token Initialization $\boldsymbol{\omega}_i^{0}$}
\label{app:init:omega}

The parameter token sequence $\boldsymbol{\omega}_i^{0} \in \mathbb{R}^{n_i \times d}$ at each layer $i$ seeds the parameter-token rollout of Eq.~\ref{eq:method-refinement-step}. For each layer $i$, a sequence of $n_i$ learnable query tokens $\boldsymbol{Q}_i \in \mathbb{R}^{n_i \times d}$ cross-attends to $\mathbf{z}_\zeta$:
\begin{equation}
\label{eq:app-init-omega}
\boldsymbol{\omega}_i^{0} \;=\; \mathrm{CrossAttn}_i\big(\boldsymbol{Q}_i, \mathbf{z}_\zeta, \mathbf{z}_\zeta\big) \;\in\; \mathbb{R}^{n_i \times d}.
\end{equation}
The query tokens supply the row count $n_i$ — the number of output neurons of layer $i$ — and absorb task-specific content from $\mathbf{z}_\zeta$ through the keys and values. Each layer has its own cross-attention module and its own query tokens, so layer-specific initialization patterns are stored in the queries and the cross-attention's projections, while task-specific information enters uniformly through $\mathbf{z}_\zeta$.

This design contrasts with HyPoGen~\citep{ren2025hypogen}, which initializes all signals from a shared, signal-independent parameter vector $\theta^{0}$ and relies on the rollout alone to introduce signal dependence; tying $\boldsymbol{\omega}_i^{0}$ to $\mathbf{z}_\zeta$ at iteration zero shortens the rollout the hypernetwork must perform from a random starting point to a task-conditioned one.

\subsection{Adaptive Scaling for Initialization-Sensitive Target Networks}
\label{app:method:adaptive-scaling}

The hypernetwork $\mathcal{H}_\Omega$ controls the target-INR weights end-to-end: at every iteration $t$, including $t=0$, the layer-$i$ weight matrix is the decoded parameter token sequence as specified in Eq.~\ref{eq:method-hypernetwork}, and the target INR is evaluated on whatever weights the hypernetwork produces. For target architectures whose stability is robust to the absolute scale of their weights, this end-to-end control raises no further issue. For target architectures with strict initialization requirements, however, the hypernetwork must additionally guarantee that the $t=0$ weight matrix $W_i^{0}$ lies within the architecture's safe range, since a $W_i^{0}$ that falls outside this range produces saturated activations or vanishing gradients at the very first iteration and prevents the rollout from being trained at all.

SIREN~\citep{sitzmann2019siren} is the canonical such case. SIREN replaces the standard non-linearity $\sigma$ in Eq.~\ref{eq:method-fwd} with the periodic activation $\sin(\Omega \cdot)$, where $\Omega$ is a frequency hyperparameter, and is sensitive to weight initialization in a way standard ReLU MLPs are not: hidden-layer weights are required to lie within the bound $\sqrt{6 / n_{i-1}} \,/\, \Omega$, which keeps the sine pre-activations within their non-saturated regime and the gradients well-conditioned through depth. The per-layer decoder of Sec.~\ref{sec:method:tokenization} maps each parameter token to its corresponding weight row, a representation-level transformation that is shared across target architectures and is independent of any one architecture's initialization scheme. Architecture-specific constraints such as SIREN's bound are therefore handled by a separate sub-module, dedicated to bridging the decoded weights and the target architecture's safe range.

Concretely, we introduce a per-layer adaptive scaling sub-module for SIREN-style targets. A learnable per-element scaling matrix $S_W^{(i)} \in \mathbb{R}^{n_i \times n_{i-1}}$ is initialized from $\mathcal{U}(-\beta_i, +\beta_i)$ with the bound
\begin{equation}
\label{eq:method-scaling-bound}
\beta_i \;=\; s_\zeta(i) \cdot \frac{\sqrt{6 / n_{i-1}}}{\Omega},
\qquad
s_\zeta(i) \;=\; \sqrt{\frac{\mathrm{var}_{\mathrm{target}}(i)}{\mathrm{var}\big(W_i^{0}\big)}},
\end{equation}
where $s_\zeta(i)$ is a variance-matching factor that aligns the empirical variance of the decoded weight $W_i^{0}$ with the target variance $\mathrm{var}_{\mathrm{target}}(i)$ implied by SIREN's safe bound, deriving the multiplier from the target architecture's own initialization scheme rather than fixing it by hand. The effective layer-$i$ weight matrix that the target INR is evaluated with becomes
\begin{equation}
\label{eq:method-scaled-weights}
\widetilde{W}_i \;=\; S_W^{(i)} \odot W_i,
\end{equation}
with $\odot$ the element-wise product, and $S_W^{(i)}$ is optimized jointly with the rest of $\mathcal{H}_\Omega$. The scaling sub-module is required only for target architectures with strict initialization constraints; for target networks without such constraints it can be omitted, and the hypernetwork's end-to-end control of the decoded weights is then sufficient on its own.

\paragraph{Necessity of adaptive scaling.}
We verify that the adaptive scaling sub-module is necessary on the two SIREN-based settings of our experiments by removing $S_W^{(i)}$ and retraining $\mathcal{H}_\Omega$ end-to-end with all other components held fixed. Reconstruction PSNR drops from $27.7$ to $11.8$ on CelebA-HQ and from $49.3$ to $8.6$ on ERA5, the latter a complete collapse: without the scaling correction the decoded $W_i^{0}$ falls outside the SIREN-safe regime at iteration zero, the periodic activations saturate, and the rollout receives no useful training signal.

\paragraph{Adaptive scaling restores trainability of LDMI across widths.}
A complementary line of evidence comes from porting our adaptive scaling sub-module onto a different hypernetwork that targets SIREN INRs. Using the width-scaling setting of Fig.~\ref{fig:width-scaling} (CelebA-HQ at batch size 64, target-INR width swept from 128 to 1024), we compare the original LDMI~\citep{peis2025hypertransforming} with an LDMI variant in which $S_W^{(i)}$ from Eq.~\ref{eq:method-scaled-weights} is inserted in front of each decoded SIREN weight matrix and trained jointly with the rest of the network, all other components held fixed. Original LDMI trains successfully at widths $128$ and $768$, but its decoded $W_i^{0}$ falls outside the SIREN-safe regime at widths $384$, $512$, and $1024$ and the run collapses, the same failure mode observed when removing $S_W^{(i)}$ from our own hypernetwork. With our adaptive scaling sub-module attached, LDMI trains stably at every width in the sweep, and PSNR rises monotonically from $20.4$ at width $128$ to $26.1$ at width $1024$. The result confirms that the bound of Eq.~\ref{eq:method-scaling-bound} is not specific to our hypernetwork architecture: it addresses an initialization constraint imposed by the SIREN target INR that any hypernetwork decoding SIREN weights must satisfy, and once the constraint is satisfied the unstable widths become trainable.

\begin{table}[h]
\centering
\small
\caption{Width-scaling stability of LDMI with and without our adaptive scaling sub-module on CelebA-HQ at batch size 64, with SIREN as the target INR. PSNR (dB) is reported; ``--'' marks runs that diverge at iteration zero because the decoded $W_i^{0}$ exits the SIREN-safe regime. The original LDMI fails on three of the five widths in the sweep, while LDMI augmented with $S_W^{(i)}$ from Eq.~\ref{eq:method-scaled-weights} trains across the full range.}
\label{tab:adaptive-scaling-ldmi}
\setlength{\tabcolsep}{6pt}
\begin{tabular}{lccccc}
\toprule
Target-INR width & 128 & 384 & 512 & 768 & 1024 \\
\midrule
LDMI (original)            & 22.1 & --   & --   & 25.3 & --   \\
LDMI + Adaptive Scaling    & 20.4 & 22.8 & 25.2 & 25.3 & 26.1 \\
\bottomrule
\end{tabular}
\end{table}

\subsection{Step-Size Sub-Module}
\label{app:method:step-size}

The refinement update of Eq.~\ref{eq:method-refinement-step} multiplies the parameter-gradient tokens by a scalar step size $\lambda_i^{t}$, which the main text introduces as a sub-module of the hypernetwork that takes $\mathbf{z}_\zeta$ as input. This appendix specifies that sub-module. We drop the iteration superscript $t$; the construction repeats identically at every iteration.

Eq.~\ref{eq:method-refinement-step} unrolls test-time optimization into $T$ feed-forward steps, but unlike standard gradient descent the hypernetwork is amortized over signals: different $\zeta$ traverse different optimization trajectories, and the step size that works for one signal does not work for another (high-frequency signals require smaller steps, smoother signals can take larger steps). A fixed scalar would underfit some signals and oscillate on others. We therefore predict $\lambda$ per-signal and per-iteration, from inputs that are available at the corresponding point in the rollout.

By analogy with the classical picture, a good step size depends on the geometry of the task (loss-surface curvature, anisotropy) and the current iterate's location in parameter space. We supply the sub-module with one input for each: $\mathbf{z}_\zeta$ encodes the per-signal conditioning that determines the task geometry, and $\boldsymbol{\omega}_i$ encodes the current parameter-token state of the layer being updated.

The sub-module concatenates $\mathbf{z}_\zeta$ and $\boldsymbol{\omega}_i$ along the token axis, applies self-attention to the joint sequence, and reduces the output to a scalar through two MLPs:
\begin{equation}
\label{eq:app-step-size}
\lambda_i \;=\; \mathrm{MLP}_{\lambda,2}\!\Big(\mathrm{vec}\big(\mathrm{MLP}_{\lambda,1}(\boldsymbol{a}_i)\big)\Big),
\qquad
\boldsymbol{a}_i \;=\; \mathrm{SelfAttn}\!\big([\mathbf{z}_\zeta;\, \boldsymbol{\omega}_i]\big).
\end{equation}
Self-attention lets every $\mathbf{z}_\zeta$ token attend to every $\boldsymbol{\omega}_i$ token and vice versa, so the predicted step size is informed jointly by the conditioning signal and the current iterate.

\subsection{Experimental Details}
\label{app:Implementation Details-new}

\subsubsection{Baseline Methods}
\label{subsubsec:baseline methods}

We compare our approach against six baselines, which we organize
into two groups based on whether their hypernetwork architecture
is optimization-oriented (Sec.~\ref{sec:method:goal}). HyPoGen is
the only prior work whose architecture mirrors the structure of
test-time optimization, and our method belongs to the same group.
The remaining five baselines---LDMI, Trans-INR, GINR-IPC, VaMoH,
and HyperDiffusion---generate INR parameters without such mirroring,
through single-pass mappings, modulation, variational inference,
or weight-space diffusion.

\paragraph{Optimization-oriented baseline.}

\textbf{HyPoGen~\citep{ren2025hypogen}} is the immediate predecessor
of our method and the only prior work whose hypernetwork architecture
is explicitly biased toward an optimization process. Originally
proposed for generalizable policy generation in robot manipulation,
HyPoGen synthesizes target-network parameters from a task
specification by approximating an optimization process executed
over a finite number of steps in a compressed latent space, with
each step instantiated by MLP sub-modules that follow the chain-rule
structure of the forward and backward computation. We adopt its
optimization-oriented principle as our starting point and replace
the MLP body with attention to overcome the quartic-in-width
scalability barrier identified in Sec.~\ref{sec:method:goal}.

\paragraph{Non-optimization-oriented baselines.}

\textbf{LDMI~\citep{peis2025hypertransforming}} extends latent diffusion models to INR generation
by replacing the standard convolutional decoder with a
Transformer-based hyper-decoder that produces INR parameters from
latent variables in a single-pass mapping. As a strong attention-based
hypernetwork without an optimization-oriented design, LDMI is the
direct counterpart for isolating the contribution of the
optimization-oriented principle in our method.

\textbf{Trans-INR~\citep{Chen2022TransformersAM}} uses a Transformer
as a meta-learner to produce the entire set of INR weights from
input observations through a set-to-set mapping. Although its design
is motivated by gradient-based meta-learning, the resulting
architecture predicts the parameters in a single forward pass without
explicitly mirroring the steps of test-time optimization. We include
it as an early Transformer-based hypernetwork baseline on the
Pre-trained NeRF setting.

\textbf{GINR-IPC~\citep{Kim2022GeneralizableIN}} takes a different approach to
amortization: instead of generating the full INR weights, it
modulates only the weights of an early MLP layer of a coordinate-based
network as an instance-specific pattern, while the remaining layers
learn shared pattern-composition rules across instances. By
restricting weight generation to a low-dimensional modulation,
GINR-IPC achieves strong generalization across images, audio, and
3D objects, and we include it as a representative modulation-based
hypernetwork baseline on Pre-trained NeRF.

\textbf{VaMoH~\citep{koyuncu2023variational}} couples INR-based variational autoencoders with
a mixture of hypernetworks, using a normalizing flow as the prior
over the latent space, in order to learn distributions over
continuous functions. We include it as a representative variational
baseline for function-space generative modeling on the 2D datasets
and ShapeNet Chairs.

\textbf{HyperDiffusion~\citep{Erkoc_2023_ICCV}} casts INR
generation as a generative-modeling problem in weight space. It
first overfits a separate MLP to each data instance to obtain a
dataset of pre-trained INR weights, and then trains a Transformer
diffusion model directly on these weights to model their
distribution. The original method is unconditional and was
demonstrated on 3D shapes and 4D mesh animations. We include
HyperDiffusion as a baseline only on the Pre-trained
NeRF~\citep{Ramirez2023DeepLO} dataset, because it is the only
dataset in our experimental setup that provides a collection of
pre-trained INR weights---one individually optimized NeRF per
object---which is a prerequisite for HyperDiffusion's weight-space
diffusion process. The other four datasets (CelebA-HQ, ImageNet,
ERA5, and ShapeNet Chairs) only provide raw signals (images,
climate fields, or occupancy grids), and obtaining a per-instance
INR weight dataset on them would require running the very
test-time optimization procedure that hypernetwork-based INR
generation aims to amortize away in the first place, defeating the
purpose of the comparison.
To enable a fair comparison under our setup, we adapt HyperDiffusion
to the conditional setting by employing a DINO
encoder~\citep{Caron2021EmergingPI} to provide image-based
conditioning signals, and instantiate it under two of our three
supervision settings as follows. Under \emph{weights only}
supervision, we recover HyperDiffusion's original training scheme:
the diffusion model is trained to denoise toward the ground-truth
NeRF weights. Under \emph{weights \& images} supervision, we
additionally render novel views from the generated weights using
the target NeRF and add an image reconstruction loss against the
ground-truth views, on top of the original weight-supervised
diffusion objective. The \emph{images only} supervision setting is
fundamentally inapplicable to HyperDiffusion: its training
objective requires ground-truth weights as the diffusion target,
and removing them leaves the diffusion process without any signal
to define what the model should denoise toward. We therefore
report ``N/A'' for HyperDiffusion under the images-only setting in
Table~\ref{tab:hyperinr-3d}.

\subsubsection{Implementation Details}
\paragraph{Training Configuration}
All experiments were conducted on NVIDIA H800 GPUs with mixed precision training. We employ the Adam optimizer with $\beta_1 = 0.9$, $\beta_2 = 0.999$, and a fixed learning rate of $5 \times 10^{-5}$ across all datasets without learning rate scheduling. 

\paragraph{Architecture Specifications}
Our hypernetwork architecture scales with dataset complexity, as shown in Table~\ref{tab:model_configs}.  The task tokenizer dimension and hidden dimensions are adjusted based on the signal complexity: simpler datasets like ShapeNet Chairs use 64-dimensional representations, while complex datasets like Pre-trained NeRF require 768 dimensions. 

\begin{table*}[htbp]
    \centering
    \caption{Model hyperparameter across different datasets.}
    \vspace{5pt}
    \footnotesize
    \setlength{\tabcolsep}{4pt}
    \label{tab:model_configs}
    \begin{tabular}{lccccc}
        \toprule
        & {CelebA-HQ 64$\times$64} & {ImageNet} & {ERA5} & {Chairs} & {Pre-trained NeRF} \\
        \midrule
        Learning rate & 5e-5 & 5e-5 & 5e-5 & 5e-5 & 5e-5 \\
        Optimizer & Adam & Adam & Adam & Adam & Adam \\
        Mini-batch size & 64 & 128 & 128 & 32 & 96 \\
        Task Tokenizer dim & 128 & 256 & 128 & 64 & 768 \\
        Each Block Hidden dim & 128 & 768 & 128 & 64 & 768 \\
        Transformer layers & 1 & 1 & 1 & 1 & 1 \\
        Token len & 384,257,257,257,4 & 384,257,257,257,4 & 342,257,257,257,2 & 171,129,2 & 64,64,64,16 \\
        \midrule
        INR type & SIREN & SIREN & SIREN & SIREN & NeRF \\
        INR layers & 5 & 5 & 5 & 3 & 3 \\
        INR hidden dim & 256 & 256 & 256 & 128 & 64 \\
        INR $\Omega$ / encode frequency & 30 & 30 & 30 & 30 & 24 \\
        \bottomrule
    \end{tabular}
\end{table*}

\subsubsection{Encoder Architectures}
\label{App:encoder architectures}

For each dataset, the encoder maps an input signal $\zeta$ to its
conditioning input $\mathbf{z}_\zeta$, from which the hypernetwork
generates the target INR parameters $\theta_\zeta$. The encoder is
shared across all compared methods on the same dataset. We follow the
encoder choices of LDMI~\citep{peis2025hypertransforming}: a 2D
convolutional VAE for CelebA-HQ and ERA5, a pre-trained and frozen
VQ-VAE~\citep{vandenOord2017NeuralDR} for ImageNet, and a 3D
convolutional VAE for ShapeNet Chairs. For Pre-trained NeRF, we use
a DINO ViT-B/16 encoder~\citep{Caron2021EmergingPI} with per-view
camera-pose conditioning injected through DiT-style adaLN-Zero
modulation~\citep{Peebles2022ScalableDM}; the encoder is initialized
from the public self-supervised DINO checkpoint and fine-tuned
end-to-end on $N{=}6$ conditional views per object, whose token
sequences are concatenated to form the conditioning input. Per-dataset specifications are
summarized in Table~\ref{tab:encoders}.

\begin{table}[h]
\centering
\small
\caption{Encoder used for each dataset.}
\label{tab:encoders}
\setlength{\tabcolsep}{8pt}
\renewcommand{\arraystretch}{1.2}
\begin{tabular}{ll}
\toprule
Dataset & Encoder \\
\midrule
CelebA-HQ        & 2D Conv VAE \\
ERA5             & 2D Conv VAE \\
ImageNet         & VQ-VAE (pre-trained, frozen) \\
ShapeNet Chairs  & 3D Conv VAE \\
Pre-trained NeRF & DINO ViT-B/16 + adaLN-Zero pose cond. \\
\bottomrule
\end{tabular}
\end{table}

\subsubsection{Loss Function}
\label{App:loss function}

For CelebA-HQ, ImageNet, ERA5, and ShapeNet Chairs, we follow the
training losses of LDMI. For Pre-trained NeRF, we use $L_1$ losses
on images and weights.

\paragraph{Notation.}
$N$ denotes the number of samples per mini-batch (for losses computed
over a patch-discriminator's output, $N$ is the total number of patch
positions across the batch). The hyperparameters
$\lambda_{\text{kl}}$, $\lambda_{\text{disc}}$,
$\lambda_{\text{codebook}}$, $\lambda_{\text{perceptual}}$,
$\lambda_{\text{weight}}$, and $\lambda_{\text{img}}$ scale the
corresponding loss terms. The
reconstruction loss $\mathcal{L}_{\text{rec}}$ defined here is the
per-instance loss $\mathcal{L}$ used in
Section~\ref{sec:method:derivation} for the chain-rule derivation; the
dataset-level training objective $\mathcal{L}_{\text{total}}$ is its
sum (with KL/adversarial/codebook terms) over a mini-batch. To avoid
clashes with the activation $\sigma$ used in Eq.~\ref{eq:method-fwd},
we write the likelihood variance as $s^2$ and the sigmoid function as
$\mathrm{sig}(\cdot)$.

For the GAN-trained datasets (CelebA-HQ and ImageNet), we additionally
use an \emph{adaptive discriminator weight} $\beta$ and a \emph{warmup
gate} $\gamma$ defined as
\begin{equation}
\beta = \lambda_{\text{disc}} \cdot \mathrm{clip}\!\left(
\frac{\|\nabla_{\theta_L}\mathcal{L}_{\text{rec}}\|}
     {\|\nabla_{\theta_L}\mathcal{L}_{\text{g}}\|+10^{-4}},
\,0,\,10^{4}\right),
\qquad
\gamma = \mathbf{1}\!\left[\mathrm{global\_step} \ge \mathrm{threshold}\right],
\end{equation}
where $\theta_L$ denotes the parameters of the last decoder layer
\citep{Esser2020TamingTF}. We write
$\mathrm{Weight}_{\text{adj}} := \gamma\cdot\beta$ for the combined
weight applied to the generator's adversarial loss; the discriminator
is gated by $\gamma$ alone.

\paragraph{CelebA-HQ.}
For the generator,
\begin{equation}
\mathcal{L}_{\text{total}}
= \mathcal{L}_{\text{rec}}
+ \lambda_{\text{kl}}\cdot\mathcal{L}_{\text{kl}}
+ \mathrm{Weight}_{\text{adj}}\cdot\mathcal{L}_{\text{g}}^{\text{hinge}}.
\end{equation}
The reconstruction loss is a per-pixel weighted Gaussian negative
log-likelihood (with learned variance) whose numerator combines pixel
and perceptual terms,
\begin{equation}
\mathcal{L}_{\text{rec}}
= \frac{1}{N}\sum_{i=1}^{N}\alpha_i\cdot
\left(\frac{|x_i-\hat{x}_i|+\lambda_{\text{perceptual}}\cdot
\mathrm{LPIPS}(x_i,\hat{x}_i)}{\exp(\log s^2)}+\log s^2\right),
\end{equation}
between the real input image $x_i$ and the reconstruction $\hat{x}_i$,
where $s^2$ is the (learned) likelihood variance and $\alpha_i$ is the
per-sample weight. The KL divergence loss is
\begin{equation}
\mathcal{L}_{\text{kl}} = \frac{1}{N}\sum_{i=1}^{N}
\mathrm{KL}\!\left(q(z_i)\,\|\,p(z_i)\right),
\end{equation}
between the posterior $q(z_i)$ and prior $p(z_i)$ of the latent
$z_i$. The generator hinge loss is
\begin{equation}
\mathcal{L}_{\text{g}}^{\text{hinge}}
= -\frac{1}{N}\sum_{i=1}^{N}\mathrm{logits\_fake}_i,
\end{equation}
where $\mathrm{logits\_fake}_i$ is the discriminator's pre-sigmoid
logit on the reconstruction $\hat{x}_i$. For the discriminator,
\begin{equation}
\mathcal{L}_{\text{discriminator}}
= \gamma\cdot\mathcal{L}_{\text{d}}^{\text{hinge}},
\qquad
\mathcal{L}_{\text{d}}^{\text{hinge}}
= \tfrac{1}{2}\bigl(\mathcal{L}_{\text{real}}+\mathcal{L}_{\text{fake}}\bigr),
\end{equation}
with
$\mathcal{L}_{\text{real}}=\frac{1}{N}\sum_{i=1}^{N}
\max(0,1-\mathrm{logits\_real}_i)$ and
$\mathcal{L}_{\text{fake}}=\frac{1}{N}\sum_{i=1}^{N}
\max(0,1+\mathrm{logits\_fake}_i)$, and
$\mathrm{logits\_real}_i$, $\mathrm{logits\_fake}_i$ the
discriminator's pre-sigmoid logits on $x_i$ and $\hat{x}_i$.
The discriminator update is gated only by the warmup indicator
$\gamma$; neither $\lambda_{\text{disc}}$ nor $\beta$ is applied here.

\paragraph{ImageNet.}
The total loss comprises an autoencoder loss and a discriminator loss.
The autoencoder loss combines reconstruction, codebook, and adversarial
terms,
\begin{equation}
\mathcal{L}_{\text{autoencoder}}
= \mathcal{L}_{\text{rec}}
+ \lambda_{\text{codebook}}\cdot\mathcal{L}_{\text{codebook}}
+ \mathrm{Weight}_{\text{adj}}\cdot\mathcal{L}_{\text{g}}^{\text{hinge}}.
\end{equation}
The reconstruction loss combines pixel-wise and perceptual terms,
\begin{equation}
\mathcal{L}_{\text{rec}}
= \mathcal{L}_{\text{pixel}}
+ \lambda_{\text{perceptual}}\cdot\mathcal{L}_{\text{perceptual}},
\end{equation}
with $\mathcal{L}_{\text{pixel}}=\|x-y\|_1$ the $L_1$ pixel loss
between the original image $x$ and the reconstruction $y$, and
$\mathcal{L}_{\text{perceptual}}=\mathrm{LPIPS}(x,y)$ the LPIPS
distance. The codebook loss regularizes the discrete latent space,
\begin{equation}
\mathcal{L}_{\text{codebook}} = \|\mathbf{z}_e-\mathbf{e}\|_2^2,
\end{equation}
where $\mathbf{z}_e$ is the encoder's continuous latent and
$\mathbf{e}$ is its nearest codebook entry. The generator hinge loss
$\mathcal{L}_{\text{g}}^{\text{hinge}}$ is defined as in the CelebA-HQ
paragraph above. The discriminator loss is
\begin{equation}
\mathcal{L}_{\text{discriminator}}
= \gamma\cdot\mathcal{L}_{\text{d}}^{\text{hinge}},
\end{equation}
with the hinge form
$\mathcal{L}_{\text{d}}^{\text{hinge}}=\tfrac{1}{2}(\mathcal{L}_{\text{real}}+\mathcal{L}_{\text{fake}})$
identical to the CelebA-HQ definition. As in the CelebA-HQ case, the
discriminator update is gated only by $\gamma$.

\paragraph{ERA5.}
\begin{equation}
\mathcal{L}_{\text{total}}
= \mathcal{L}_{\text{rec}}+\lambda_{\text{kl}}\cdot\mathcal{L}_{\text{kl}},
\qquad
\mathcal{L}_{\text{rec}}
= \frac{1}{N}\sum_{i=1}^{N}
\left(\frac{|x_i-\hat{x}_i|}{\exp(\log s^2)}+\log s^2\right),
\end{equation}
where $x_i$ and $\hat{x}_i$ are the original and reconstructed
temperature fields, $s^2$ is the (learned) likelihood variance, and
$\mathcal{L}_{\text{kl}}$ is the KL divergence defined above. ERA5
uses neither per-sample weighting nor a perceptual term.

\paragraph{ShapeNet Chairs.}
\begin{equation}
\mathcal{L}_{\text{total}}
= \mathcal{L}_{\text{rec}}+\lambda_{\text{kl}}\cdot\mathcal{L}_{\text{kl}},
\end{equation}
with the per-voxel binary cross-entropy reconstruction loss
\begin{equation}
\mathcal{L}_{\text{rec}}
= -\frac{1}{N}\sum_{i=1}^{N}\alpha_i\cdot
\left(x_i\log\mathrm{sig}(\ell_i)
+(1-x_i)\log\bigl(1-\mathrm{sig}(\ell_i)\bigr)\right),
\end{equation}
where $x_i\in\{0,1\}$ is the binary occupancy of voxel $i$, $\ell_i$
is the predicted logit, and $\mathrm{sig}(\cdot)$ is the sigmoid
function. The factor $\alpha_i$ is an optional per-voxel weight that
can be supplied to compensate for the imbalance between occupied and
empty voxels; in our default ShapeNet configuration no such weight is
passed and $\alpha_i\equiv 1$.

\paragraph{Pre-trained NeRF.}
\begin{equation}
\mathcal{L}_{\text{total}}
= \lambda_{\text{weight}}\cdot\mathcal{L}_{\text{weight}}
+ \lambda_{\text{img}}\cdot\mathcal{L}_{\text{img}}.
\end{equation}
The weight loss is the $L_1$ distance between the predicted and
ground-truth NeRF parameters,
\begin{equation}
\mathcal{L}_{\text{weight}}
= \frac{1}{N}\sum_{i=1}^{N}
\bigl\|\theta_i^{\text{pred}}-\theta_i^{\text{gt}}\bigr\|_1,
\end{equation}
and the image loss is the $L_1$ distance between rendered and
ground-truth novel-view RGB,
\begin{equation}
\mathcal{L}_{\text{img}}
= \frac{1}{N}\sum_{i=1}^{N}
\bigl\|\hat{\mathbf{c}}_i-\mathbf{c}_i^{\text{gt}}\bigr\|_1,
\end{equation}
where $\theta_i^{\text{pred}}$ and $\theta_i^{\text{gt}}$ are the
hypernetwork-predicted and ground-truth NeRF weight vectors of the
$i$-th object, and $\hat{\mathbf{c}}_i$ and $\mathbf{c}_i^{\text{gt}}$
are the rendered and ground-truth RGB at the corresponding novel
views. For weight-only supervision,
$\lambda_{\text{weight}}=1$ and $\lambda_{\text{img}}=0$; for
image-only supervision, $\lambda_{\text{weight}}=0$ and
$\lambda_{\text{img}}=1$; for combined supervision,
$\lambda_{\text{weight}}=0.1$ and $\lambda_{\text{img}}=1$.

\subsubsection{Additional Qualitative results}
\label{app:additional-qualitative-results}
\begin{figure*}[!t] 
\centering
\includegraphics[width=1.0\linewidth]{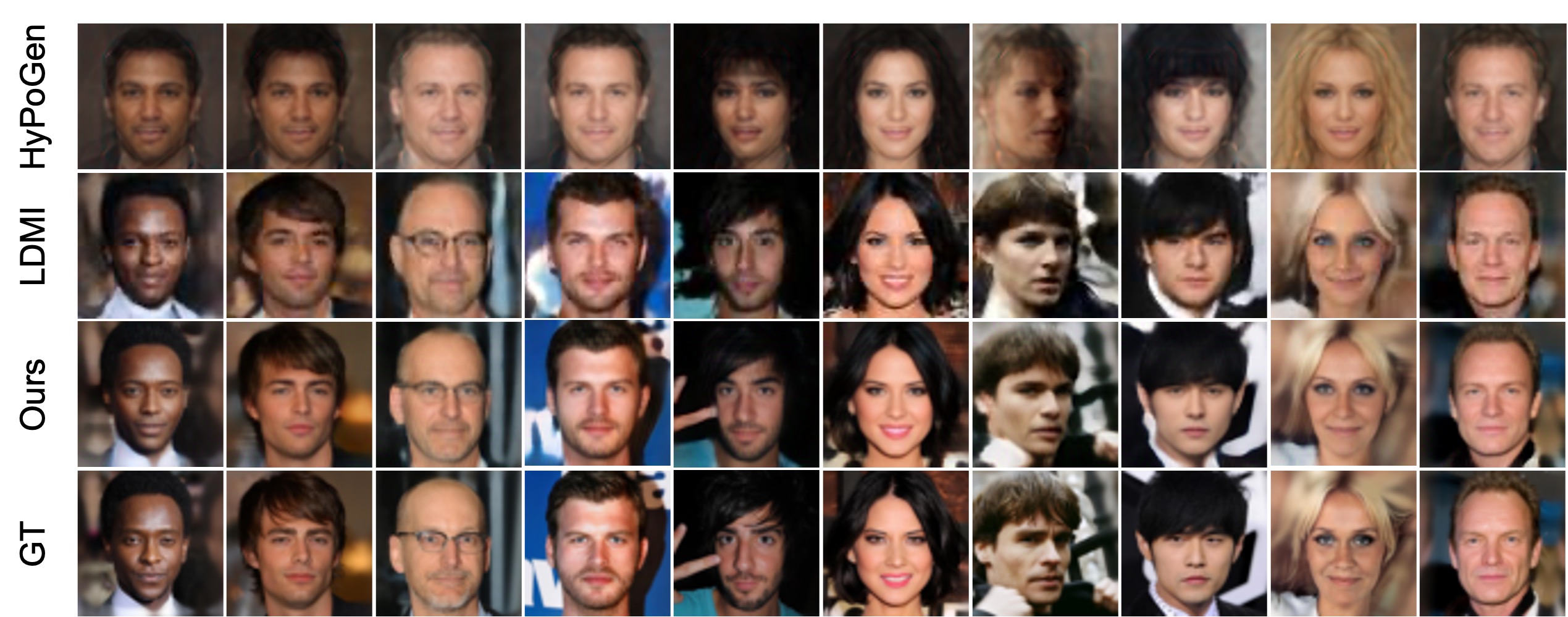}
\caption{Additional visual comparison on CelebA-HQ.}
\label{fig:app_celebahq}
\end{figure*}

\begin{figure*}[!t] 
\centering
\includegraphics[width=1.0\linewidth]{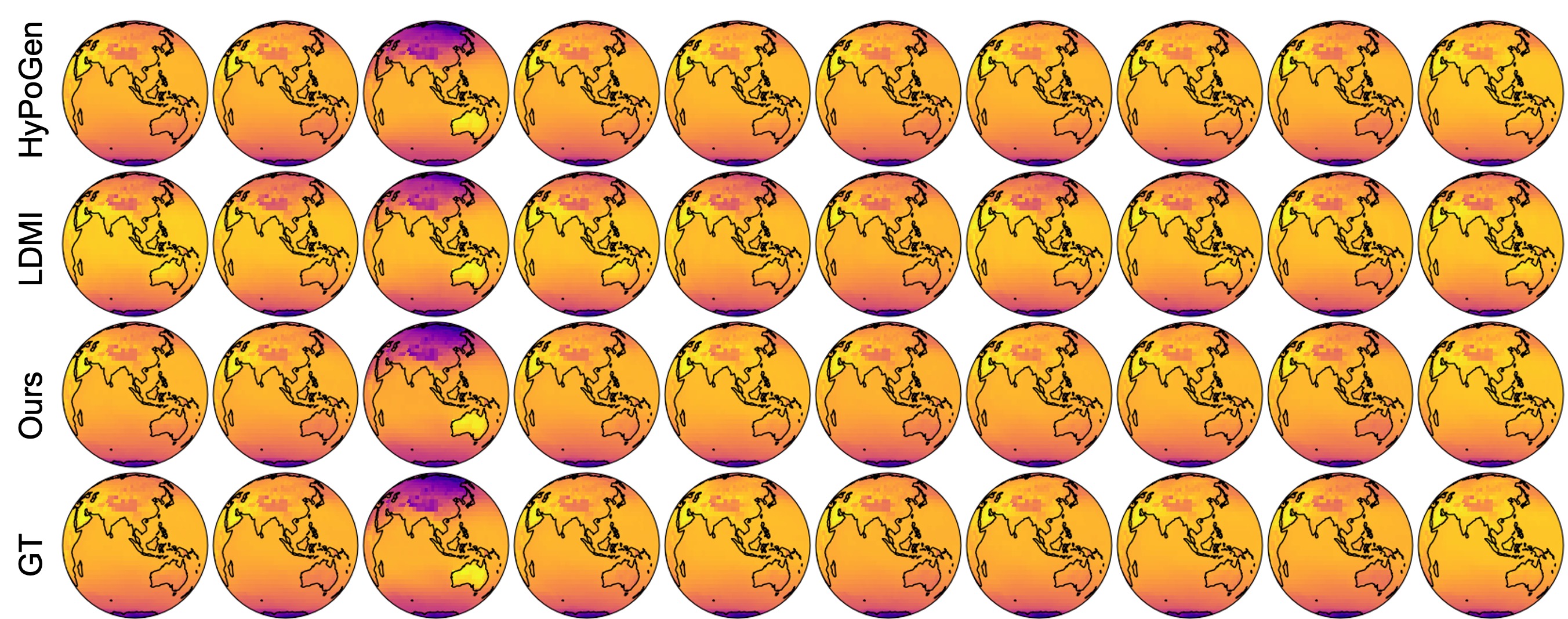}
\caption{Additional visual comparison on ERA5.}
\label{fig:app_era5}
\end{figure*}

\begin{figure*}[!t] 
\centering
\includegraphics[width=1.0\linewidth]{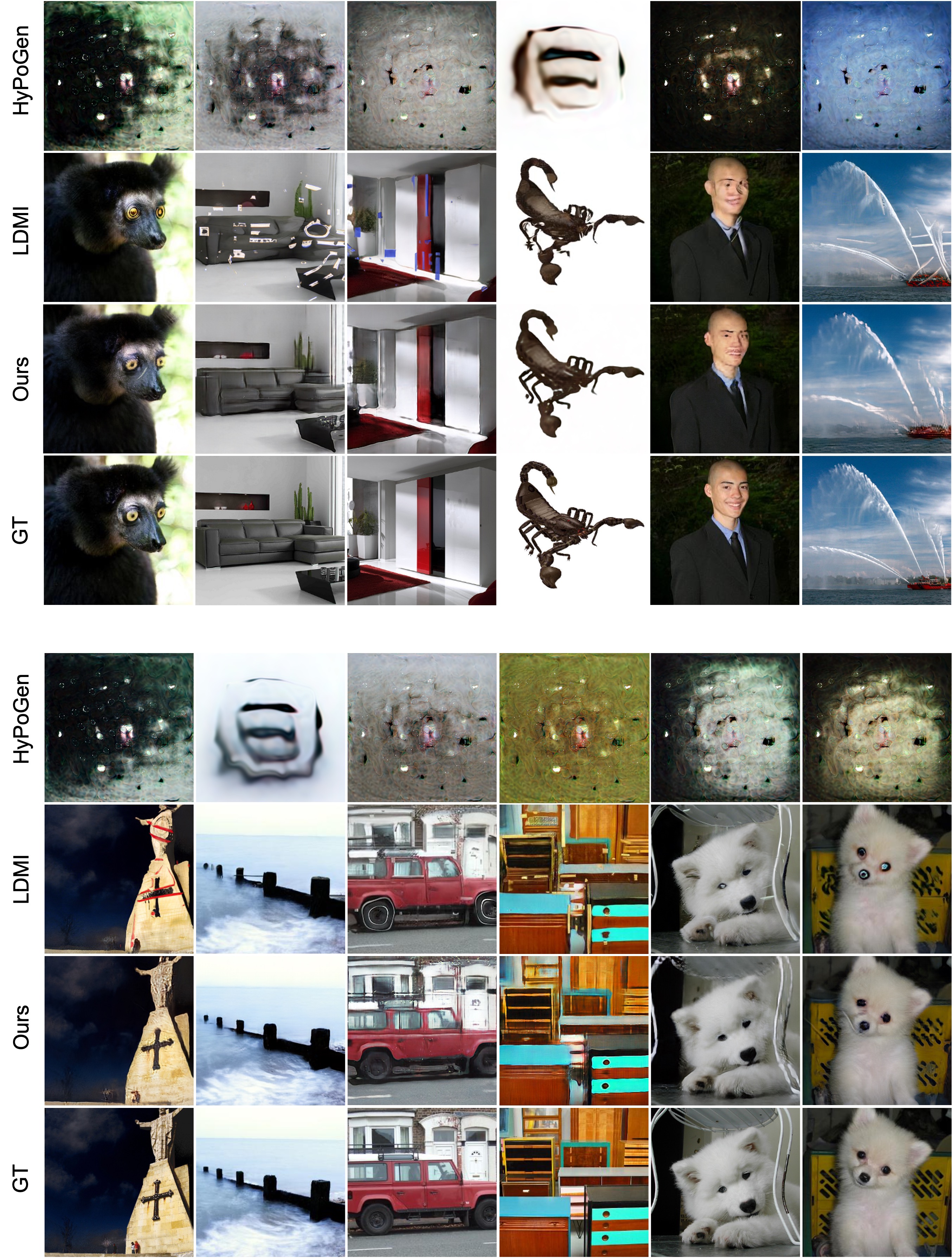}
\caption{Additional visual comparison on ImageNet.}
\label{fig:app_imagenet}
\end{figure*}

\begin{figure*}[!t] 
\centering
\includegraphics[width=0.9\linewidth]{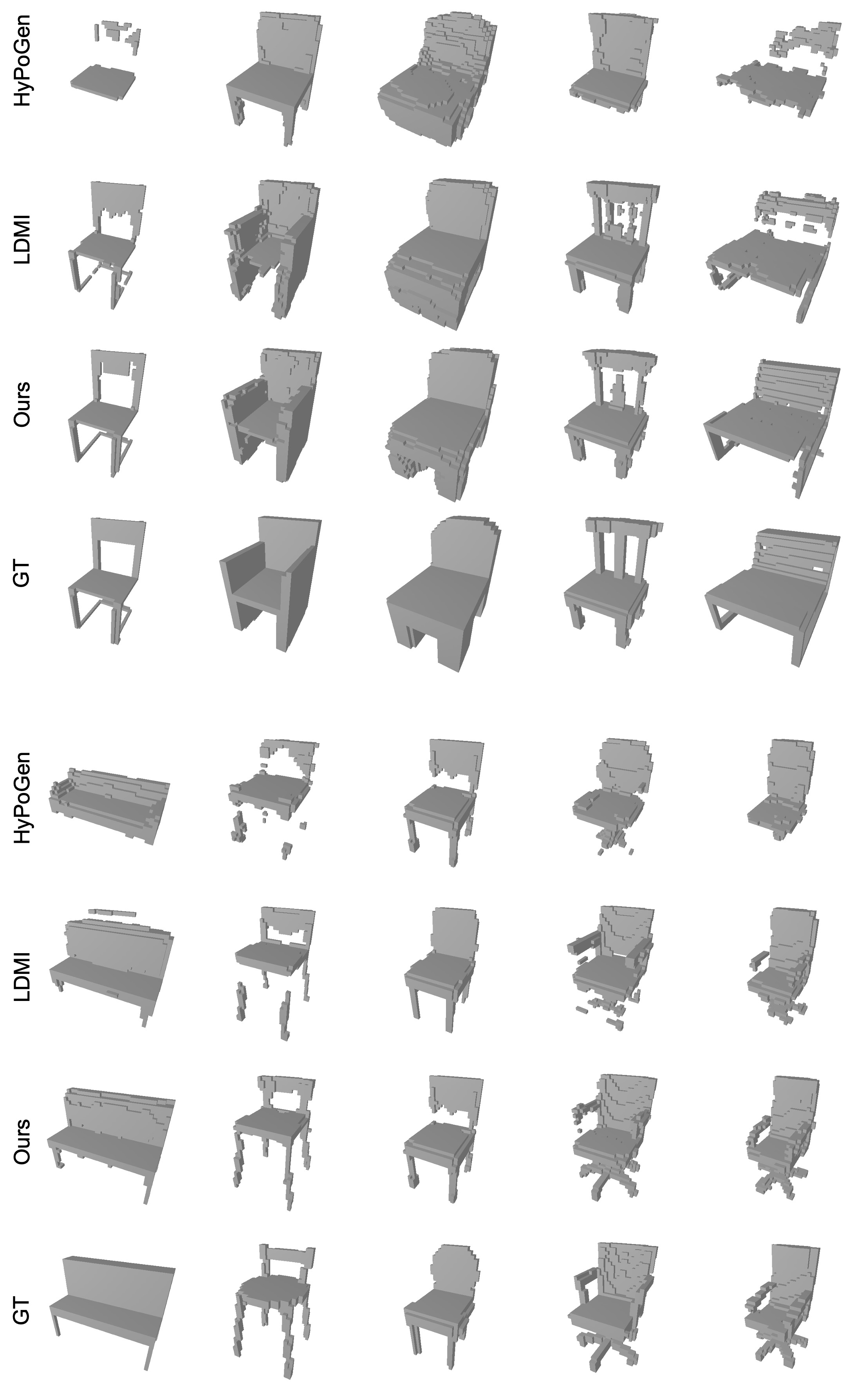}
\caption{Additional visual comparison on ShapeNet Chairs.}
\label{fig:app_shapenet}
\end{figure*}

\begin{figure*}[!t] 
\centering
\includegraphics[width=0.92\linewidth]{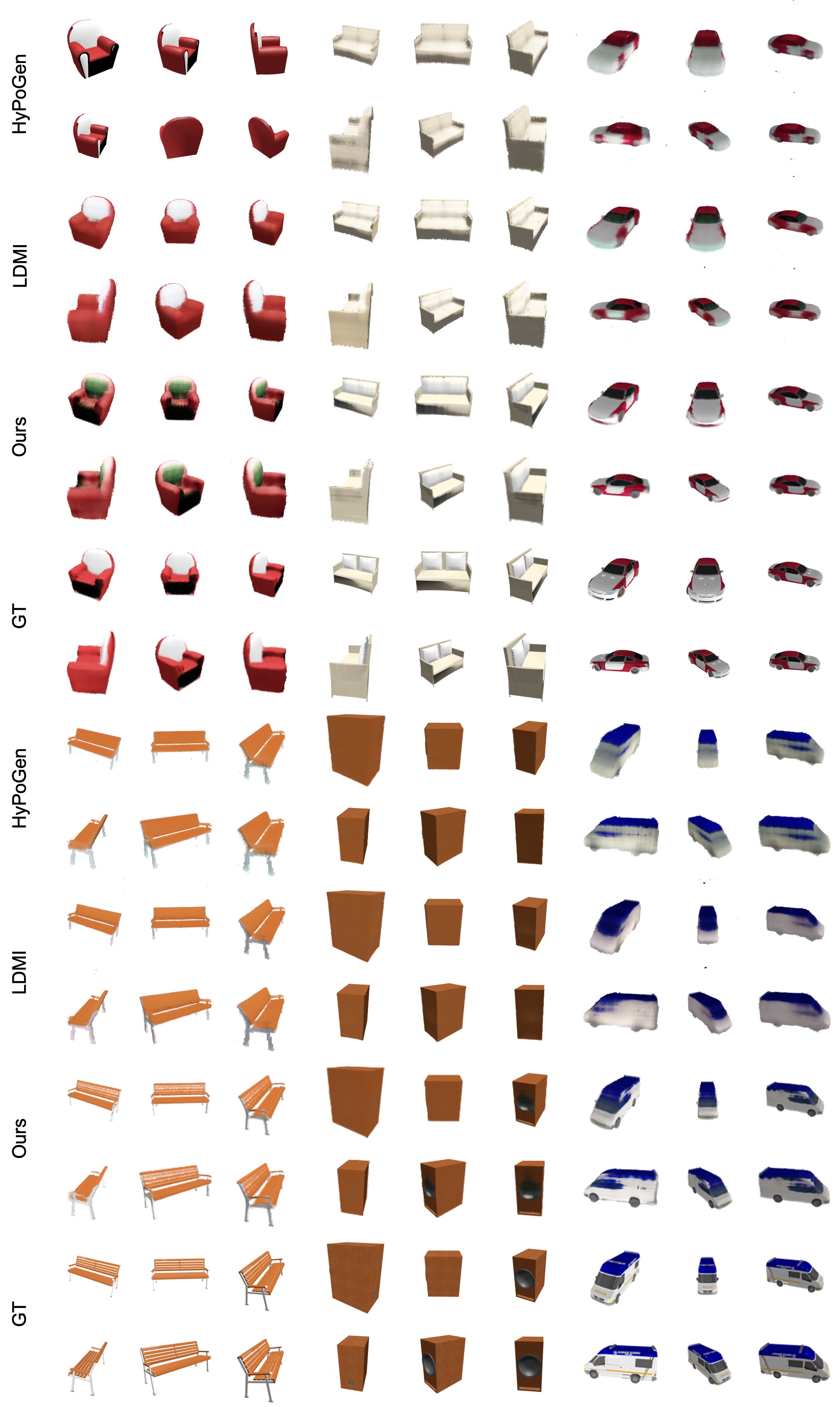}
\caption{Additional visual comparison on Pre-trained NeRF.}
\label{fig:app_nerf}
\end{figure*}

We provide additional visual comparisons on all five datasets:
CelebA-HQ (Figure~\ref{fig:app_celebahq}),
ERA5 (Figure~\ref{fig:app_era5}),
ImageNet (Figure~\ref{fig:app_imagenet}),
ShapeNet Chairs (Figure~\ref{fig:app_shapenet}), and
Pre-trained NeRF (Figure~\ref{fig:app_nerf}).
In each figure, rows from top to bottom correspond to
HyPoGen, LDMI, Ours, and the ground truth (GT), while
columns show different samples.
Across all datasets, our method produces reconstructions
that most closely match the ground truth.

\subsubsection{Additional Quantitative Results}
\label{app:additional-quantitative-results}

\paragraph{NeRF-specific baselines.}
We further compare against two NeRF-dedicated approaches,
Trans-INR~\citep{Chen2022TransformersAM} and
GINR-IPC~\citep{Kim2022GeneralizableIN}, on the Pre-trained NeRF
dataset (Table~\ref{tab:nerf-extra-baselines}). Our method
outperforms both across all three supervision settings.

\begin{table}[tb]
    \centering
    \caption{Additional NeRF-specific baselines on the Pre-trained
    NeRF dataset under three supervision settings. Our method
    outperforms both Trans-INR and GINR-IPC across all settings.}
    \label{tab:nerf-extra-baselines}
    \small
    \setlength{\tabcolsep}{6pt}
    \begin{tabular}{llccc}
        \toprule
        Supervision & Metric & Trans-INR & GINR-IPC & \textbf{Ours} \\
        \midrule
        Weights only
            & PSNR $\uparrow$    & 16.5   & 17.1   & \textbf{22.0}   \\
        \midrule
        Images only
            & PSNR $\uparrow$    & 21.9   & 26.1   & \textbf{28.8}   \\
        \midrule
        \multirow{3}{*}{Weights \& Images}
            & PSNR $\uparrow$    & 16.6   & 18.2   & \textbf{27.8}   \\
            & SSIM $\uparrow$    & 0.8659 & 0.8899 & \textbf{0.948}  \\
            & LPIPS $\downarrow$ & 0.2618 & 0.1855 & \textbf{0.0742} \\
        \bottomrule
    \end{tabular}
\end{table}

\paragraph{LPIPS on ImageNet.}
For completeness, we report LPIPS on ImageNet, which was omitted
from Table~\ref{tab:hyperinr-2d} due to space constraints:
VAMoH 0.4128, HyPoGen 0.6141, LDMI \textbf{0.1468}, and Ours
0.1636. LDMI achieves slightly lower LPIPS than our method on
this dataset, while our method consistently outperforms all
baselines on PSNR and SSIM.

\subsection{Additional Ablations}
\label{app:more-ablations}

\paragraph{Forward chain.}
\textbf{w/o Forward Chain} removes the recursive application of
$\phi_i$ along $i = 1, \dots, L$ and replaces the pseudo-activation
tokens $\boldsymbol{\tau}_i$ with learnable token sequences that do
not depend on $\boldsymbol{\tau}_{i-1}$, while keeping the backward
block, the rollout depth $T$, and the parameter count comparable
(Table~\ref{tab:arch-ablation}). PSNR drops by 2.8 / 4.4 / 3.2\,dB.
The drop is the cost of removing the token-space analogue of
Eq.~\ref{eq:method-fwd}: when $\boldsymbol{\tau}_i$ no longer carries
the layer-wise activation history that $\phi_i$ would have produced
from $\boldsymbol{\tau}_{i-1}$ and $\boldsymbol{\omega}_i$, the
backward block has to compute Jacobians and parameter gradients
against tokens that no longer represent the activations they are
meant to surrogate.

\paragraph{Backward block decomposition.}
\textbf{w/o Backward Structure} replaces the four role-specific
backward modules of Section~\ref{sec:method:iterative-module}
($\psi^{\boldsymbol{\tau}}_i$, $\psi^{\boldsymbol{\omega}}_i$,
$\psi^{\boldsymbol{\delta}}_{i-1}$, and $\psi^{\nabla}_i$) with
four identical $\Delta\boldsymbol{\omega}$-generating cross-attentions
stacked sequentially, matched in parameter count
(Table~\ref{tab:arch-ablation}). PSNR drops by 3.3 / 4.8 / 3.6\,dB,
showing that what drives the backward block's contribution is not
raw attention capacity but the chain-rule decomposition of one
optimization step into the four quantities of
Eqs.~\ref{eq:method-Jtau}--\ref{eq:method-delta-grad}. Removing the
decomposition removes the structure that the optimization-oriented
principle requires.

\paragraph{Number of refinement steps $T$.}
\textbf{Single-Step Rollout ($T{=}1$)} reduces the iterative module
of Section~\ref{sec:method:rollout} from three refinement steps to
one, so the hypernetwork emits the target weights after a single
forward--backward pass over the target network rather than three.
PSNR drops by 0.3 / 0.8 / 1.1\,dB on CelebA-HQ / ImageNet / NeRF
(Table~\ref{tab:iterative-module-1}), with the drop growing with the
size and diversity of the dataset. The pattern is consistent with
the role of $T$ in the rollout of Eq.~\ref{eq:method-refinement-step}:
each refinement step applies one parameter-token update, and a single
step is enough on simpler datasets but leaves residual error on
harder ones, where the cumulative effect of multiple updates becomes
necessary to reach the final weights. The improvements from $T{=}1$
to $T{=}3$ are smaller than the gaps caused by the structural
ablations of Section~\ref{sec:analysis}, which we view as further
evidence that the structural choices, rather than the rollout depth,
are what drive our method's accuracy.

\begin{table}[h]
\centering
\small
\setlength{\tabcolsep}{4pt} 
\caption{Ablations on the structural blocks of the iterative module of Sec.~\ref{sec:method:iterative-module}. 
Each variant disables one structural block---the forward chain, the role-specific backward decomposition, while keeping all other components and total compute matched to the full model. 
PSNR ($\uparrow$) on CelebA-HQ, ImageNet, and Pre-trained NeRF (\emph{Weights\,\&\,Images}).}
\begin{tabular}{l cc cc cc}
\toprule
& \multicolumn{2}{c}{\textbf{CelebA-HQ}} & \multicolumn{2}{c}{\textbf{ImageNet}} & \multicolumn{2}{c}{\textbf{NeRF}} \\
\cmidrule(lr){2-3} \cmidrule(lr){4-5} \cmidrule(lr){6-7}
\textbf{Variant} & Params & PSNR & Params & PSNR & Params & PSNR \\
\midrule
Full model           & 41.2 & 27.7 & 124.5 & 22.9 & 482.5 & 27.8 \\
w/o Forward module   & 33.1 & 24.9 & 101.2 & 18.5 & 441.9 & 24.6 \\
w/o Backward module  & 38.4 & 24.4 & 110.1 & 18.1 & 456.8 & 24.2 \\
\bottomrule
\end{tabular}
\label{tab:arch-ablation}
\end{table}

\begin{table}[h]
\centering
\small
\setlength{\tabcolsep}{5pt}
\caption{Ablations on the structural blocks of the iterative module of Sec.~\ref{sec:method:iterative-module}. 
PSNR ($\uparrow$) on CelebA-HQ, ImageNet, and Pre-trained NeRF (\emph{Weights\,\&\,Images}).}
\label{tab:iterative-module-1}
\begin{tabular}{l|ccc}
\toprule
\textbf{Variant} & CelebA-HQ & ImageNet & NeRF \\
\midrule
\textbf{Ours (Full, $T{=}3$)}     & \textbf{27.7} & \textbf{22.9} & \textbf{27.8} \\
\midrule
Single Layer Optimization ($T{=}1$) & 27.4 & 22.1 & 26.7 \\
\bottomrule
\end{tabular}
\end{table}

\subsection{Stability Across Random Seeds}
\label{app:stability}
 
To assess robustness with respect to random initialization, we evaluate our
method across multiple random seeds on three datasets that span the diversity
of our experimental setup: Pre-trained NeRF (3D radiance fields), CelebA-HQ
(2D natural images), and ERA5 (2D climate data).
Table~\ref{tab:psnr_seeds_all} reports PSNR for five different seeds, along
with mean and standard deviation. The variability captured corresponds to
random-initialization differences while keeping all other experimental
conditions fixed.
 
Our method exhibits strong consistency across all three datasets. On
Pre-trained NeRF, PSNR remains within $27.7$--$27.8$~dB
(mean: $27.76 \pm 0.05$). On CelebA-HQ, the standard deviation is
$0.10$~dB. On ERA5, PSNR ranges from $48.9$ to $49.5$~dB
(mean: $49.26 \pm 0.23$). The observed standard deviations
($0.05$--$0.23$~dB) are an order of magnitude smaller than the gaps
between our method and the baselines reported in
Tables~\ref{tab:hyperinr-2d} and~\ref{tab:hyperinr-3d}, indicating that the
improvements are not attributable to seed variance.
 
\begin{table}[ht]
\caption{PSNR stability across random seeds on multiple datasets. Each row
reports our method's PSNR under five independent training runs that differ
only in the random seed used for initialization, together with the mean and
standard deviation. NeRF results are reported under the \emph{Weights\,\&\,Images} supervision setting.}
\vspace{5pt}
\centering
\small
\setlength{\tabcolsep}{4pt}
\begin{tabular}{l|ccccc|c}
\toprule
Dataset & Seed 444 & Seed 555 & Seed 777 & Seed 888 & Seed 999 & Mean $\pm$ Std \\
\midrule
Pre-trained NeRF
& 27.7 & 27.8 & 27.7 & 27.8 & 27.8
& 27.76 $\pm$ 0.05 \\
CelebA-HQ
& 27.8 & 27.7 & 27.7 & 27.5 & 27.6
& 27.66 $\pm$ 0.10 \\
ERA5
& 48.9 & 49.3 & 49.4 & 49.2 & 49.5
& 49.26 $\pm$ 0.23 \\
\bottomrule
\end{tabular}
\label{tab:psnr_seeds_all}
\end{table}

\subsection{Engineering Strategies for Scalability}
\label{app:engineering-strategies}
 
We explore two simple strategies to mitigate the depth-linear and
attention-quadratic cost factors of Appendix~\ref{app:limitations}.
\textbf{\emph{Token Compression}} ($n{=}2$) merges every $n$ adjacent
parameter tokens after the row-wise tokenization of
Sec.~\ref{sec:method:tokenization} into a single token, halving the
attention sequence length (from $n_i$ to $n_i / n$); the target
network's width is unchanged, but each cross-attention module operates
on a shorter sequence, addressing wider target networks.
\textbf{\emph{Layer Fusion}} ($m{=}2$) groups every $m$ consecutive
target-network layers and lets each group share a single set of
forward/backward modules (the five cross-attentions of
Sec.~\ref{sec:method:iterative-module}); the target network's depth is
unchanged, but the hypernetwork now instantiates only
$\lceil L/m \rceil$ rather than $L$ such module sets, addressing
deeper target networks. We evaluate both strategies on ImageNet and
Pre-trained NeRF in Table~\ref{tab:tricks2}.
 
On ImageNet, all three configurations share the same target network
($5{\times}256$ SIREN), so the comparison is direct: Token Compression
and Layer Fusion reduce inference time at batch size~$16$ by $8\%$
($58.5\to54.1$~ms) and $30\%$ ($58.5\to41.2$~ms) respectively, Layer
Fusion also cuts parameter count by $20\%$ ($124.5$M~$\to99.4$M), and
PSNR drops by less than $0.3$~dB.
 
On Pre-trained NeRF (under the \emph{Images only} supervision setting),
the two variants intentionally use \emph{larger} target networks than
the baseline (Token Comp.\ $3{\times}128$, Layer Fus.\ $6{\times}64$),
so PSNR is not directly comparable ($\dagger$). The relevant question
is how much these larger targets inflate the hypernetwork's cost:
inference time rises by only $4$--$6\%$ and parameter count by under
$2\%$, while PSNR stays within $0.4$~dB of the baseline. Together
with the ImageNet result, this shows the two strategies either
reduce cost at fixed target size or absorb larger targets at
near-flat cost.
 
\begin{table*}[t]
\centering
\small
\caption{
Efficiency comparison (batch size: 16) under two practical engineering
strategies for scaling to wider or deeper target networks.
NeRF results are reported under the \emph{Images only} supervision setting.
}
\begin{tabular}{lcccccc}
\toprule
& \multicolumn{3}{c}{\textbf{ImageNet}} & \multicolumn{3}{c}{\textbf{Pre-trained NeRF}} \\
\cmidrule(lr){2-4} \cmidrule(lr){5-7}
\textbf{Metric} & Token Comp. & Layer Fus. & Ours & Token Comp. & Layer Fus. & Ours \\
\midrule
PSNR (dB)                 & 22.7   & 22.6  & 22.9  & 29.0$^\dagger$ & 29.2$^\dagger$ & 28.8 \\
Infer Time @ b=16 (ms)   & 54.1    & 41.2   & 58.5   & 2098           & 2135           & 2018 \\
Params (M)                & 132.82 & 99.43 & 124.5 & 484.39         & 491.43         & 482.5 \\
\bottomrule
\end{tabular}
\vspace{2pt}
\begin{flushleft}
\footnotesize{$^\dagger$ Uses a larger target network than the baseline,
so PSNR is not directly comparable.
For ImageNet, all methods use the same target network ($5{\times}256$).
For Pre-trained NeRF, the two variants use larger target networks:
Token Comp.\ uses $3{\times}128$, Layer Fus.\ uses $6{\times}64$.}
\end{flushleft}
\label{tab:tricks2}
\end{table*}

\subsection{Limitations}
\label{app:limitations}
We discuss limitations of our method along three axes.
 
\paragraph{Per-instance inference latency.}
Our forward and backward blocks are realized as stacks of cross-attention modules,
each of which carries non-trivial constant-factor cost relative to a single MLP
layer. Although Table~\ref{tab:batch16} shows that our method matches or
improves on HyPoGen and LDMI in parameter count and GPU memory at batch size~$16$,
single-instance inference (batch size~$1$) carries higher latency than the
MLP-based HyPoGen on the smaller benchmarks.
 
\paragraph{Depth and width scaling of the iterative module.}
Our parameter-count derivation in Appendix~\ref{app:param-count}
establishes that the cost of one refinement step is independent of
target-network width, but it is linear in the number of target-network
layers~$L$, and the attention sub-modules are quadratic in their
sequence length. For target networks substantially deeper or wider than
the $3$--$5$~layer SIRENs and NeRFs evaluated in this work, the
depth-linear and attention-quadratic factors would dominate the
iterative module's cost. We discuss two simple engineering strategies
that mitigate this cost in Appendix~\ref{app:engineering-strategies}.
  
\paragraph{Scope of statistical evaluation.}
We report multi-seed stability on three of the five datasets
(Appendix~\ref{app:stability}). Running the full multi-seed protocol on all
five datasets and all six baselines was beyond our compute budget,
and we therefore cannot report mean$\,\pm\,$std for every cell of
Tables~\ref{tab:hyperinr-2d}--\ref{tab:hyperinr-3d}. The seed-to-seed
standard deviations we do measure ($0.05$--$0.23$~dB) are an order of
magnitude smaller than the gaps between our method and the baselines,
so we expect this restriction not to affect our conclusions, but a fully
seeded benchmark across all configurations remains future work.

\subsection{Asset Licenses and Attribution}
\label{app:licenses}
 
We use the following datasets and baseline implementations. All assets are
used in accordance with their respective licenses and terms of use.
 
\paragraph{Datasets.}
\begin{itemize}
\setlength{\itemsep}{2pt}
\item \textbf{CelebA-HQ}~\citep{karras2018progressive}, derived from CelebA~\citep{liu2015deep}.
  Released for non-commercial research use under the original CelebA terms.
\item \textbf{ImageNet}~\citep{russakovsky2015imagenet}. Released for
  non-commercial research use under the ImageNet Terms of Access.
\item \textbf{ERA5}~\citep{hersbach2019era5}. Released by the Copernicus
  Climate Change Service (C3S) under the Copernicus Licence; downloaded
  from the Copernicus Climate Data Store.
\item \textbf{ShapeNet Chairs}~\citep{Chang2015ShapeNetAI}. Released for
  non-commercial research use under the ShapeNet Terms of Use.
\item \textbf{Pre-trained NeRF dataset}~\citep{Ramirez2023DeepLO}. Used under
  the license stated by the original release.
\end{itemize}
 
\paragraph{Baseline implementations.}
We compare with HyPoGen~\citep{ren2025hypogen},
LDMI~\citep{peis2025hypertransforming},
VaMoH~\citep{koyuncu2023variational},
Trans-INR~\citep{Chen2022TransformersAM},
GINR-IPC~\citep{Kim2022GeneralizableIN}, and
HyperDiffusion~\citep{Erkoc_2023_ICCV}. We use the official implementations
where available, under the licenses specified by their authors. Encoder
backbones for the conditioning networks of
Appendix~\ref{App:encoder architectures} are used
under their respective public licenses.

\subsection{Broader Impacts}
\label{app:impact}
 
This research develops a hypernetwork architecture for generating Implicit
Neural Representations and does not involve human subjects, animal
experiments, or sensitive data collection. All experiments use publicly
available benchmark datasets (CelebA-HQ, ImageNet, ERA5, ShapeNet Chairs,
Pre-trained NeRF) under their original licenses; no privacy or ethical
issues arise from data usage.
 
The proposed methodology advances fundamental research in neural
representation learning. We see two classes of broader impact worth noting.
 
\paragraph{Positive impacts.}
Our method reduces the parameter count of the hypernetwork's core body
relative to MLP-based predecessors by a factor that scales with target-network
width (Appendix~\ref{app:param-count}), and reduces inference-time GPU memory
on the largest benchmark by $5$--$6\times$ (Table~\ref{tab:batch16}). For
downstream applications such as 3D reconstruction, neural rendering, and
scientific signal modeling that rely on hypernetwork-generated INRs, this
lowers the compute and memory barrier and makes such tools more accessible
to researchers and practitioners with limited hardware.
 
\paragraph{Potential negative impacts.}
As with most generative-modeling research, improvements in neural-field
generation could in principle be combined with other systems to produce
synthetic visual content. Our work focuses on the architecture for parameter
generation rather than on photorealistic media synthesis, and we do not
foresee a direct path from our contribution to harmful applications. The
CelebA-HQ dataset contains images of human faces; we use it only as a
reconstruction-quality benchmark and do not train any identification,
attribution, or generation model targeting specific individuals.
 
\subsection{Use of Large Language Models}
\label{app:llm}
 
Large language models were used in this work only as auxiliary writing aids,
specifically for surface-level revisions to phrasing and grammar. They were
not used to design the method, derive results, run experiments, or generate
any scientific content. All ideas, derivations, experimental designs, and
conclusions are the authors' own intellectual contributions, and the use of
LLMs does not affect the methodology, scientific rigor, or originality of
the research.

\end{document}